\documentclass{article}

\usepackage[nonatbib,preprint]{neurips_2025}

\usepackage{amsmath,amsfonts,bm}

\def\eqref#1{equation~\ref{#1}}

\def\1{\bm{1}}

\DeclareMathAlphabet{\mathsfit}{\encodingdefault}{\sfdefault}{m}{sl}
\SetMathAlphabet{\mathsfit}{bold}{\encodingdefault}{\sfdefault}{bx}{n}

\DeclareMathOperator*{\argmax}{arg\,max}

\usepackage{amssymb,bm,amsthm}
\usepackage{subfig}
\usepackage{floatrow}
\usepackage{multirow}
\usepackage{array}
\usepackage{color}
\usepackage{colortbl}
\usepackage{wrapfig}

\usepackage{graphicx}
\usepackage{bbding}
\usepackage[utf8]{inputenc} 
\usepackage[T1]{fontenc}    
\usepackage{url}            
\usepackage{booktabs}       
\usepackage{amsfonts}       
\usepackage{nicefrac}       
\usepackage{xcolor}         
\usepackage{microtype}

\usepackage{nccmath}

\usepackage{amsmath}
\usepackage{mathtools}
\definecolor{deepgreen}{rgb}{ .439,  .678,  .278}
\definecolor{blue}{rgb}{ .357,  .608,  .835}
\definecolor{orange}{rgb}{ .929,  .49,  .192}

\newcommand\etal{\textit{et al.}}
\newcommand\ie{\textit{i.e.,}}
\newcommand\eg{\textit{e.g.,}}

\newcommand\wrt{\textit{w.r.t.}}

\newcommand{\red}{\textcolor{red}}

\newcommand{\beq}{\begin{equation}}
\newcommand{\eeq}{\end{equation}}
\newcommand{\beqnn}{\begin{equation*}}
\newcommand{\eeqnn}{\end{equation*}}
\newcommand{\beqy}{\begin{eqnarray}}
\newcommand{\eeqy}{\end{eqnarray}}
\newcommand{\beqynn}{\begin{eqnarray*}}
\newcommand{\eeqynn}{\end{eqnarray*}}
\newcommand{\bit}{\begin{itemize}}
\newcommand{\eit}{\end{itemize}}
\newcommand{\ben}{\begin{enumerate}}
\newcommand{\een}{\end{enumerate}}
\newcommand{\bex}{\begin{example}}
\newcommand{\eex}{\end{example}}

\newcommand{\balg}[1]{\begin{algorithm} \caption{#1}}
\newcommand{\ealg}{\end{algorithm}}

\newcommand{\balgc}{\begin{algorithmic}[1]}
\newcommand{\ealgc}{\end{algorithmic}}

\newcommand{\bary}{\begin{array}}
\newcommand{\eary}{\end{array}}
\newcommand{\bmx}{\begin{bmatrix}}
\newcommand{\emx}{\end{bmatrix}}
\newcommand{\bsmx}{\left[\begin{smallmatrix}}
\newcommand{\esmx}{\end{smallmatrix}\right]}
\newcommand{\bmxc}[1]{\left[\begin{array}{@{}#1@{}}}
\newcommand{\emxc}{\end{array}\right]}
\newcommand{\bcn}{\begin{center}}
\newcommand{\ecn}{\end{center}}

\newcommand{\Rbb}{{\mathbb{R}}}

\usepackage[numbers,compress,sort]{natbib}	

\title{Re-evaluating the Advancements of Heterophilic Graph Learning}

\author{
Sitao Luan$^{2,*}$, Qincheng Lu$^{1,*}$,  Chenqing Hua$^{1,2}$,  Xinyu Wang$^1$, Jiaqi Zhu$^{1}$, Xiao-Wen Chang$^{1}$ \\
\{sitao.luan@mail, chenqing.hua@mail, qincheng.lu@mail, xinyu.wang5@mail, \\jiaqi.zhu@mail, chang@cs\}.mcgill.ca\\
$^1$McGill University; $^2$ Mila - Quebec Artificial Intelligence Institute; $^*$Equal Contribution\\
}

\begin{document}

\maketitle

\begin{abstract}
\vspace{-0.1cm}
Over the past decade, Graph Neural Networks (GNNs) have achieved great success on machine learning tasks with relational data. However, recent studies have found that heterophily can cause significant performance degradation of GNNs, especially on node-level tasks. Numerous heterophilic benchmark datasets have been put forward to validate the efficacy of heterophily-specific GNNs, and various homophily metrics have been designed to help recognize these challenging datasets. Nevertheless, there still exist multiple pitfalls that severely hinder the proper evaluation of new models and metrics: 1) lack of hyperparameter tuning; 2) insufficient evaluation on the truly challenging heterophilic datasets; 3) missing quantitative evaluation for homophily metrics on synthetic graphs. To overcome these challenges, we first train and fine-tune baseline models on $27$ most widely used benchmark datasets, and categorize them into three distinct groups: malignant, benign and ambiguous heterophilic datasets. We identify malignant and ambiguous heterophily as the truly challenging subsets of tasks, and to our best knowledge, we are the first to propose such taxonomy. Then, we re-evaluate $11$ state-of-the-arts (SOTA) GNNs, covering six popular methods, with fine-tuned hyperparameters on different groups of heterophilic datasets. Based on the model performance, we comprehensively reassess the effectiveness of different methods on heterophily. At last, we evaluate $11$ popular homophily metrics on synthetic graphs with three different graph generation approaches. To overcome  the unreliability of observation-based comparison and evaluation, we conduct the first quantitative evaluation and provide detailed analysis.
\end{abstract}
\vspace{-0.3cm}
\section{Introduction}
\vspace{-0.2cm}
As a generic data structure, graph is capable of modeling complex relations among objects in many real-world problems~\cite{sperduti1993encoding, goller1996learning, sperduti1997supervised, frasconi1998general, hua2024enzymeflow}. In the last decade, various Graph Neural Networks (GNNs) architectures have been proposed~\cite{scarselli2008graph, bruna2014spectral, defferrard2016convolutional, kipf2016classification, hamilton2017inductive, gilmer2017neural, velivckovic2018graph, xu2018powerful, luan2019break, hua2024reactzyme} and shown to outperform traditional neural networks (NNs) in modeling graph-based real-world tasks~\cite{monti2017geometric, ying2018graph, pfaff2020learning, zhao2021consciousness, bongini2021molecular, hua2024mudiff, lu2024gcepnet}.
\vspace{-0.2cm}

The success of GNNs, especially on node-level tasks, is commonly believed to be attributed to the homophily principle~\cite{mcpherson2001birds}, which means that connected nodes are more likely to have similar labels~\cite{pei2020geom} or attributes~\cite{hamilton2020graph}. This inductive bias is believed to be a major contributor to the superiority of GNNs over traditional NNs on various tasks~\cite{zhu2020beyond}. 
On the other hand, heterophily, \ie{} the lack of homophily~\cite{lozares2014homophily, luan2024heterophilic}, 
is often cited as the primary reason for the inferiority of GNNs on heterophilic graphs.
In such graphs, edges typically connect nodes from different classes, which can lead to mixed and indistinguishable node embeddings during the message passing process~\cite{zhu2020beyond, luan2022complete}. Recently, numerous models have been proposed to deal with heterophily~\cite{pei2020geom, zhu2020beyond, luan2022complete, bo2021beyond, lim2021new, chien2021adaptive, he2021bernnet, li2022finding, luan2022revisiting, lu2024flexible, luan2024heterophilic, zhenglet, zheng2024rethinking} and many homophily metrics have been put forward to identify the graph datasets that are unfriendly to GNNs~\cite{luan2024heterophilic,zheng2024missing}. 

However, to date, there is no work that strictly validate the recent advances in heterophilic graph learning. Upon examination, we empirically identify several pitfalls that can severely impede fair and accurate assessment of models and metrics:
\vspace{-0.15cm}
\begin{itemize}
    \item  \textbf{Lack of Adequate Hyperparameter Tuning.} With careful hyperparameter tuning, it is empirically found that basic GNNs can actually outperform some heterophily-specific models on several heterophilic graphs~\cite{ma2021homophily, luo2024classic}. This indicates that there potentially exist a substantial amount of inaccurate and biased results reported in current literature, which hinder fair  comparison and mislead our understanding of heterophily problem.
    \item \textbf{Insufficient Evaluation on Truly Challenging Heterophilic Datasets.}
   Based on the studies in~\cite{ma2021homophily,luan2021heterophily,luan2023graph}, the real challenging heterophilic datasets are those where graph-aware models underperform graph-agnostic models, instead of those with small homophily values. With this criterion, the evaluation results on many commonly used benchmark datasets cannot adequately validate the effectiveness of models  on heterophily.
    \item \textbf{Absence of Quantitative Evaluation Benchmark for Homophily Metrics.} 
    The existing evaluation methods mainly depends on observing how well the homophily metric correlates with GNN performance on synthetic graphs.
    However, such observation-based comparison can easily lead to subjective and unreliable conclusions, and there is currently no quantitative evaluation of homophily metrics.
\end{itemize}
\vspace{-0.1cm}
In this paper, we aim to address the above issues and our main contributions are:
\begin{itemize}
    \item \textbf{Discover New Categorization of Heterophily and Identify the Challenging Ones.} To find out the real challenging heterophilic datasets, in Section~\ref{sec:malignant_benigh_ambiguous_heterophily}, we fine-tune baseline graph-aware models and their corresponding graph-agnostic models on $27$ mostly used benchmark datasets.
    We find that there exist three disjoint sets of heterophilic datasets,
    where graph-aware models: 1) consistently outperform graph-aware models; 2) consistently underperform graph-aware models; 3) have inconsistent performance against graph-aware models. Based on this discovery, we categorize them into three types of heterophilic graphs: malignant, benign and ambiguous, 
    and we argue that the malignant and ambiguous datasets are the truly challenging ones which should be used to validate the effectiveness of models. 
    \item \textbf{Reassess Popular Heterophily-specific Graph Models.} In Section~\ref{sec:reassessment_sota}, we reassess $11$ state-of-the-arts (SOTA) GNNs with fine-tuned hyperparameters on each categories of the $27$ benchmark datasets. Based on the results, the efficacy of some widely used methods is found questionable, \eg{} most SOTA GNNs are not significantly better than the best baseline models in any category of heterophilic datasets, and some of them actually compromise their performance on easy graphs to obtain better results on challenging graphs. 
    \item \textbf{Quantitative Evaluation for Homophily Metrics.} In Section~\ref{sec:homophily_metric_evaluation_comparison}, we evaluate $11$ popular homophily metrics on synthetic graphs with three graph generation approaches. To compare them strictly and accurately, in Section~\ref{sec:quantitative_benchmark_metrics}, we use Pearson correlation and  Fréchet distance to assess the metrics quantitatively. We conduct detailed analysis and provide insights for future evaluation.
\end{itemize}

\vspace{-0.3cm}
\section{Preliminaries}
\vspace{-0.2cm}
\subsection{Notation}
\vspace{-0.1cm}
We define a graph $\mathcal{G}=(\mathcal{V},\mathcal{E})$, where $\mathcal{V}=\{1,2,\ldots,N\}$ is the set of nodes and $\mathcal{E}=\{e_{ij}\}$ is the set of edges without self-loops. The adjacency matrix of $\mathcal{G}$ is denoted by $A=(A_{i,j})\in \Rbb^{N\times N}$ with $A_{i,j}=1$ if there is an edge between nodes $i$ and $j$, otherwise $A_{i,j}=0$.  The diagonal degree matrix of $\mathcal{G}$ is denoted by $D$ with $D_{i,i} = d_i = \sum_j A_{i,j}$. The neighborhood set $\mathcal{N}_i$ of node $i$ is defined as $\mathcal{N}_i=\{j: e_{ij} \in \mathcal{E}\}$. A graph signal is a vector in $\mathbb{R}^N$, whose $i$-th entry is a feature of node $i$.  Additionally, we use ${X} \in \mathbb{R}^{N\times F_h}$ to denote the feature matrix, whose columns are graph signals and $i$-th row ${X_{i,:}} = \bm{x}_i^\top$ is the feature vector of node $i$  (we use \textbf{bold} font for vectors).
The label encoding matrix is $Y \in \mathbb{R}^{N\times C}$, where $C$ is the number of classes, and its $i$-th row $Y_{i,:}$ is the one-hot encoding of the label of node $i$. We denote $y_i = \argmax_{j} Y_{i,j} \in \{1,2,\dots C\}$. 
The indicator function $\bm{1}_B$ equals 1 when event $B$ happens and 0 otherwise. 

For nodes $i,j \in \mathcal{V}$, if $y_i=y_j$, they are termed \textit{intra-class nodes}; if $y_i \neq y_j$, they are termed \textit{inter-class nodes}. Similarly, an edge $e_{i,j} \in \mathcal{E}$ is termed an \textit{intra-class edge} if $y_i = y_j$, and an \textit{inter-class edge} if $y_i \neq y_j$.

The affinity matrices can be derived from the adjacency matrix, \eg{} $A_\text{rw} = D^{-1} A$ and $A_\text{sym} = D^{-1/2} A D^{-1/2}$. After applying the renormalization trick~\cite{kipf2016classification}, we have $\hat{A}_\text{sym} = \tilde{D}^{-1/2} \tilde{A} \tilde{D}^{-1/2}$ and $ \hat{A}_{\text{rw}} = \tilde{D}^{-1} \tilde{A}$, where $\tilde{A} \equiv A+I$ and $\tilde{D} \equiv D+I$. The renormalized affinity matrix essentially adds a self-loop to each node. The affinity matrices are commonly used as aggregation operators in GNNs.

\vspace{-0.2cm}
\subsection{Graph-aware Models and Graph-agnostic Models} 
\vspace{-0.1cm}
\label{sec:graph_aware_and_agnostic}
A network that incorporates feature aggregation based on graph structure is referred to as a graph-aware model~\cite{luan2023graph}, \eg{} GCN~\cite{kipf2016classification}, SGC~\cite{wu2019simplifying}; and a network that does not use graph structure information in each layer is called graph-agnostic model, such as MLP-2 (Multi-Layer Perceptron with 2 layers) and MLP-1. A graph-aware model is always coupled with a graph-agnostic model, as when the aggregation step is removed, the graph-aware model becomes exactly the same as its coupled graph-agnostic model,
\eg{} GCN is coupled with MLP-2 and SGC-1 is coupled with MLP-1 as shown below:
\begin{equation}
\begin{aligned}
    \label{eq:gcn_original}
   &\textbf{GCN: } \text{Softmax} (\hat{A}_\text{sym} \; \text{ReLU} (\hat{A}_\text{sym} {X} W_0 ) \; W_1 ),\ \ \textbf{MLP-2: } \text{Softmax} (  \text{ReLU} (  {X} W_0 ) \; W_1 ),\\
   &\textbf{SGC-1: } \text{Softmax} (\hat{A}_\text{sym}  {X} W_0 ), \; \textbf{MLP-1: } \text{Softmax} ( {X} W_0 ),
   \end{aligned}
\end{equation}
where $W_0 \in \Rbb^{F_0 \times F_1}$ and $W_1 \in \Rbb^{F_1\times O}$ are learnable parameter matrices. A node classification task on graph is considered as real challenging if a graph-aware model underperforms its coupled graph-agnostic counterpart on it~\cite{luan2023graph}. Numerous homophily metrics have been proposed to recognize the difficult graphs and the most commonly used ones will be introduced in the next subsection.

\vspace{-0.2cm}
\subsection{Homophily Metrics}
\vspace{-0.1cm}
\label{sec:metrics_on_homogeneous_graphs}
There are mainly four ways to define homophily metrics~\cite{luan2024heterophilic}. We will introduce their calculations briefly in this subsection. See a more detailed summary of the metrics in Appendix~\ref{appendix:metrics_on_homogeneous_graphs}.
\vspace{-0.1cm}
\paragraph{Graph-Label Consistency} There are four commonly used homophily metrics that are based on the consistency between node labels and graph structures, including edge homophily~\cite{abu2019mixhop,zhu2020beyond}, node homophily~\cite{pei2020geom}, class homophily~\cite{lim2021new} and adjusted homophily~\cite{platonov2023characterizing},
defined as follows:
\begin{equation}
\begin{aligned}
\label{eq:homo_metrics_graph_label_consistency}
& \text{H}_\text{edge}(\mathcal{G}) = \frac{\big|\{e_{uv} \mid e_{uv}\in \mathcal{E}, y_{u} = y_{v}\}\big|}{|\mathcal{E}|}; \ \  \text{H}_\text{node}(\mathcal{G}) = \frac{1}{|\mathcal{V}|} \sum_{v \in \mathcal{V}} \frac{\big|\{u \mid u \in \mathcal{N}_v, y_{u} = y_{v}\} \big|}{d_v}; \\
& \text{H}_\text{class}(\mathcal{G}) \!=\! \frac{1}{C\!-\!1} \sum_{k=1}^{C}\bigg[h_{k}
    \!-\! \frac{\big|\{v \!\mid\! Y_{v,k} \!=\! 1 \}\big|}{N}\bigg]_{+}, \ \ h_{k}\! =\! \frac{\sum_{v \in \mathcal{V}, Y_{v,k}  =  1} \big|\{u \!\mid\!  u \in \mathcal{N}_v, y_{u}\!=\! y_{v}\}\big| }{\sum_{v \in \{v|Y_{v,k}=1\}} d_{v}};  \\
& \text{H}_\text{adj}(\mathcal{G}) = \frac{\text{H}_\text{edge} - \sum_{c=1}^C \bar{p}_c^2}{1-\sum_{c=1}^C \bar{p}_c^2}, \ \ 
\bar{p}_c = \frac{\sum_{v: y_v=c} d_v}{2|\mathcal{E}|},  
\end{aligned}
\end{equation}
where $[a]_{+}=\max (a, 0)$, $h_{k}$ is the class-wise homophily metric~\cite{lim2021new}.
\vspace{-0.1cm}
\paragraph{Similarity-Based Metrics} Generalized edge homophily~\cite{jin2022raw} and aggregation homophily~\cite{luan2022revisiting} leverages similarity functions to define the metrics:
\begin{equation}
\begin{aligned}
\label{eq:homo_metrics_graph_similarity_based}
& \resizebox{0.94\hsize}{!}{$\text{H}_\text{GE} (\mathcal{G}) = \frac{\sum_{(i,j) \in \mathcal{E}} \cos(\bm{x}_{i}, \bm{x}_{j})}{|\mathcal{E}|};\ \ 
\text{H}_{\text{agg}}(\mathcal{G}) =  \frac{1}{\left| \mathcal{V} \right|} \times \left| \left\{v \,\big| \, \mathrm{Mean}_u  \big( \{S(\hat{{A}},Y)_{v,u}^{y_{u} =y_{v}} \}\big) \geq  \mathrm{Mean}_u\big(\{S(\hat{{A}},Y)_{v,u}^{y_{u} \neq y_{v}}  \} \big) \right\} \right|$}, \\
\end{aligned}
\end{equation}
where $\mathrm{Mean}_u\left(\{\cdot\}\right)$ takes the average over $u$ of a given multiset of values or variables and $S(\hat{{A}},Y)=\hat{{A}}Y(\hat{{A}}Y)^\top$ is the post-aggregation node similarity matrix. These two metrics are feature-dependent.
\vspace{-0.1cm}
\paragraph{Neighborhood Identifiability/Informativeness}  Label informativeness~\cite{platonov2023characterizing} and neighborhood identifiability~\cite{chen2023exploiting} use the neighbor distribution instead of pairwise comparison to define the metrics:
\begin{equation}
\begin{aligned}
\label{eq:homo_metrics_graph_identifiable_informative}
\resizebox{0.94\hsize}{!}{$ \text{LI} 
= 2 -\frac{\sum\limits_{c_1, c_2} p_{c_1, c_2 } \ln \frac{p_{c_1, c_2}}{\bar{p}_{c_1} \bar{p}_{c_2}}}{\sum_c \bar{p}_c \log \bar{p}_c};\ \  
\text{H}_{\text {neighbor }}(\mathcal{G}) = \sum\limits_{k=1}^C \frac{n_k}{N} \text{H}_{\text {neighbor }}^k, \ \ 
\text{H}_{\text {neighbor }}^k 
= \frac{-\sum\limits_{i=1}^C \tilde{\sigma}_i^k \ln \left( \tilde{\sigma}_i^k \right)}{\ln (C)}$}
\end{aligned}
\end{equation}
where $p_{c_1, c_2} = \sum_{(u, v) \in \mathcal{E}} \frac{\bm{1} \left\{y_u=c_1, y_v=c_2 \right\}}{2|\mathcal{E}|}$ for $c_1, c_2 \in \{1,\dots,C\}$; 
$n_k$ is the number of nodes with the label $k$;
and $\tilde{\sigma}_i^k$ will be defined immediately. 
Let $A^k \in \mathbb{R}^{n_k \times C}$ be a class-level neighborhood label distribution matrix for class $k=1, \ldots, C$, \ie{} for a node $i$ from class $k$, $(A^k)_{i,c}$ is the proportion of the neighbors of node $i$ belonging to class $c$,
and let $\sigma_1^k, \sigma_2^k, \ldots, \sigma_C^k$ denote the singular values of $A^k$, 
and they are normalized such that $\sum_{c=1}^C \tilde{\sigma}_c^k=1$, 
i.e., $\tilde{\sigma}_c^k= \sigma_c^k/\sum_{c=1}^C \sigma_c^k$.
\vspace{-0.1cm}
\paragraph{Hypothesis Testing Based Performance Metrics}  Classifier-based performance metric (CPM)~\cite{luan2023graph} uses the p-value of the following hypothesis testing as a metric to measure the node distinguishability of the aggregated features $H$ compared with the original features $X$.
\begin{equation}
\begin{aligned}
\label{eq:definition_homophily_metrics}
\resizebox{0.94\hsize}{!}{$\text{H}_0: \text{Acc}(\text{Classifier}(H)) \geq \text{Acc}(\text{Classifier}(X));\ \ 
\text{H}_1: \text{Acc}(\text{Classifier}(H)) < \text{Acc}(\text{Classifier}(X)) $,}
\end{aligned}
\end{equation}
where Acc is the prediction accuracy of the given classifier.  To capture the feature-based linear or non-linear information efficiently, Luan \etal~\cite{luan2023graph} choose Gaussian Naïve Bayes (GNB)~\cite{hastie2009elements} and Kernel Regression (KR) with Neural Network Gaussian Process (NNGP)~\cite{lee2018deep,arora2019exact, garriga2018deep,matthews2018gaussian} as the classifiers, which do not require iterative training.

Overall, $\text{H}_\text{adj}$ can assume negative values, while other metrics all fall within the range of $[0,1]$. Except for $\text{H}_{\text {neighbor}}(\mathcal{G})$, 
where a smaller value indicates more identifiable,\footnote{To compare with other metrics more easily, in this paper, we use $1-\text{H}_{\text {neighbor}}(\mathcal{G})$ for quantitative analysis.} the other metrics with higher values indicate strong homophily, implying that graph-aware models are more likely to outperform their coupled graph-agnostic model, and vice versa. These metrics will be compared in Section~\ref{sec:synthetic_comparison}.

\vspace{-0.3cm}
\section{Categorization of Heterophily Datasets and Model Re-Evaluation}
\vspace{-0.2cm}
\label{sec:homogeneous_benchmark}
In this section, we conduct a series of experiments with fine-tuned hyperparameters for accurate assessment and fair comparison of GNNs built for heterophilic graphs. Specifically, in Section~\ref{sec:experimental_settings}, we introduce the $27$ benchmark datasets used in this paper and the experimental setups;  in Section~\ref{sec:malignant_benigh_ambiguous_heterophily}, based on the performance of fine-tuned graph-aware and graph-agnostic models, we classify the existing heterophily benchmark datasets into malignant, benign and ambiguous groups, and we argue that the real challenging tasks are on malignant and ambiguous datasets; in Section~\ref{sec:reassessment_sota}, we re-evaluate $11$ popular SOTA models with fine-tuned hyperparameters on each group of heterophilic graphs to reassess their effectiveness and disclose their limitations on addressing heterophily.
\vspace{-0.2cm}
\subsection{Experimental Settings}
\vspace{-0.1cm}
\label{sec:experimental_settings}
We collect $27$ mostly used benchmark datasets for heterophily research~\cite{pei2020geom, rozemberczki2021multi, lim2021new, lim2021large, sun2022beyond, platonov2022critical, zhou2024opengsl}. The dataset names and data splits are, 
\begin{itemize}
    \item \textit{Cornell, Wisconsin, Texas, Film} are from~\cite{pei2020geom}, \textit{Chameleon, Squirrel} are from~\cite{rozemberczki2021multi}, \textit{Cora, CiteSeer, PubMed} are from~\cite{yang2016revisiting}. We use the data processed by~\cite{pei2020geom}. The models are trained on $10$ random splits with 60\%/20\%/20\% for train/validation/test, which follows~\cite{chien2021adaptive}.
    \item \textit{Deezer-Europe, genius, arXiv-year, Penn94, pokec, snap-patents, twitch-gamers} are from~\cite{lim2021new,lim2021large}. We train models on each dataset with five fixed 50\%/25\%/25\% splits for train/validation/test, which is the same as~\cite{lim2021new,lim2021large}.
    \item \textit{roman-empire, amazon-ratings, minesweeper, tolokers, questions, Chameleon-filtered, Squirrel-filtered} are from~\cite{platonov2022critical}. The models are trained on $10$ fixed splits with 50\%/25\%/25\% samples for train/validation/test, which is provided by~\cite{platonov2022critical}.
    \item \textit{BlogCatalog, Flickr, BGP, Wiki-cooc} are from~\cite{zhou2024opengsl, sun2022beyond}. The splits for training/validation/test are $10$ 60\%/20\%/20\% random splits, which is the same as~\cite{chien2021adaptive}.
\end{itemize}
For other experimental settings such as early stopping, optimizer, max number of training epochs, evaluation metrics, we all follow the original papers. The hyperparameter searching range is shown in Appendix \ref{appendix:hyperparameter_searching_range}.
\vspace{-0.1cm}
\paragraph{Computing Resources}
For all experiments on real-world and synthetic datasets, we use NVIDIA V100 GPUs with 16/32GB GPU memory. 
The software implementation is based on PyTorch and PyTorch Geometric \cite{fey2019fast}.
\vspace{-0.2cm}
\subsection{Categorization of Heterophily Datasets}
\vspace{-0.1cm}
\label{sec:malignant_benigh_ambiguous_heterophily}
\begin{table*}[htbp]
  \centering
  \caption{Categorization of benchmark datasets. The cells marked by \textcolor{deepgreen}{green} are the better results for the comparison of graph-aware models vs. graph-agnostic models.}
  \resizebox{1\hsize}{!}{
    \begin{tabular}{c|c|c|ccccccp{5.145em}|cccc|c}
    \toprule
    \toprule
    \multicolumn{2}{p{11.21em}|}{Categories} & {Datasets} & {\#nodes} & {\#edges} & {\#feature dim} & {\#classes} & {$\text{H}_\text{edge}$} & {$\text{H}_\text{node}$} & Eval Metric  &{GCN} &{MLP-2} & {SGC-1} & {MLP-1} & {Literature} \\
    \midrule
          &       & {Cornell} & 183   & 295   & 1,703 & 5     & 0.2983 & 0.2001 & Accuracy & 82.46 $\pm$ 3.11 & \cellcolor[rgb]{ .439,  .678,  .278}91.30 $\pm$ 0.70 & 70.98 $\pm$ 8.39 & \cellcolor[rgb]{ .439,  .678,  .278}93.77 $\pm$ 3.34 & \cite{pei2020geom} \\
          &       & {Wisconsin} & 251   & 499   & 1,703 & 5     & 0.1703 & 0.0991 & Accuracy & 75.5 $\pm$ 2.92 & \cellcolor[rgb]{ .439,  .678,  .278}93.87 $\pm$ 3.33 & 70.38 $\pm$ 2.85 & \cellcolor[rgb]{ .439,  .678,  .278}93.87 $\pm$ 3.33 & \cite{pei2020geom} \\
          &       & {Texas} & 183   & 309   & 1,703 & 5     & 0.0615 & 0.0555 & Accuracy & 83.11 $\pm$ 3.2 & \cellcolor[rgb]{ .439,  .678,  .278}92.26 $\pm$ 0.71 & 83.28 $\pm$ 5.43 & \cellcolor[rgb]{ .439,  .678,  .278}93.77 $\pm$ 3.34 & \cite{pei2020geom} \\
          &       & {Film} & 7,600 & 33,544 & 931   & 5     & 0.2163 & 0.2023 & Accuracy & 35.51 $\pm$ 0.99 & \cellcolor[rgb]{ .439,  .678,  .278}38.58 $\pm$ 0.25 & 25.26 $\pm$ 1.18 & \cellcolor[rgb]{ .439,  .678,  .278}34.53 $\pm$ 1.48 & \cite{pei2020geom} \\
          & {Malignant} & {Deezer-Europe} & 28,281 & 92,752 & 31,241 & 2     & 0.5251 & 0.5299 & Accuracy & 62.23 $\pm$ 0.53 & \cellcolor[rgb]{ .439,  .678,  .278}66.55 $\pm$ 0.72 & 61.63 $\pm$ 0.25 & \cellcolor[rgb]{ .439,  .678,  .278}63.14 $\pm$ 0.41 & \cite{lim2021new} \\
          &       &  genius & 421,961 & 984,979 & 12    & 2     & 0.6176 & 0.0985 & Accuracy & 83.26 $\pm$ 0.14 & \cellcolor[rgb]{ .439,  .678,  .278}86.62 $\pm$ 0.08 & 82.31 $\pm$ 0.45 & \cellcolor[rgb]{ .439,  .678,  .278}86.48 $\pm$ 0.11 & \cite{lim2021large} \\
          &       &  roman-empire  & 22,662 & 32,927 & 300   & 18    & 0.0469 & 0.046 & Accuracy & 48.92 $\pm$ 0.46 & \cellcolor[rgb]{ .439,  .678,  .278}66.04 $\pm$ 0.71 & 44.60 $\pm$ 0.52 & \cellcolor[rgb]{ .439,  .678,  .278}64.12 $\pm$ 0.61 & \cite{platonov2022critical} \\
          &       & BlogCatalog & 5,196 & 171,743 & 8,189 & 6     & 0.4011 & 0.3914 & {Accuracy} & 79.67 $\pm$ 1.06 & \cellcolor[rgb]{ .439,  .678,  .278}92.97 $\pm$ 0.89 & 71.07 $\pm$ 1.15 & \cellcolor[rgb]{ .439,  .678,  .278}91.86 $\pm$ 0.93 & \cite{zhou2024opengsl} \\
          &       & Flickr & 7,575 & 239,738 & 12,047 & 9     & 0.2386 & 0.2434 & {Accuracy} & 71.38 $\pm$ 1.00 & \cellcolor[rgb]{ .439,  .678,  .278}90.24 $\pm$ 0.96 & 60.10 $\pm$ 1.21 & \cellcolor[rgb]{ .439,  .678,  .278}89.91 $\pm$ 0.97 & \cite{zhou2024opengsl} \\
          &       & BGP   & 63,977 & 174,803 & 287   & 8     & 0.2545 & 0.083 & {Accuracy} & 62.56 $\pm$ 0.94 & \cellcolor[rgb]{ .439,  .678,  .278}65.56 $\pm$ 0.55 & 61.74 $\pm$ 0.73 & \cellcolor[rgb]{ .439,  .678,  .278}64.67 $\pm$ 0.81 & \cite{sun2022beyond} \\
\cmidrule{2-15}    {Heterophily } &       & {Chameleon} & 2,277 & 36,101 & 2,325 & 5     & 0.2339 & 0.2467 & Accuracy & \cellcolor[rgb]{ .439,  .678,  .278}64.18 $\pm$ 2.62 & 46.72 $\pm$ 0.46 & \cellcolor[rgb]{ .439,  .678,  .278}64.86 $\pm$ 1.81 & 45.01 $\pm$ 1.58 & \cite{rozemberczki2021multi} \\
    {Graphs } &       & {Squirrel} & 5,201 & 217,073 & 2,089 & 5     & 0.2234 & 0.2154 & Accuracy & \cellcolor[rgb]{ .439,  .678,  .278}44.76 $\pm$ 1.39 & 31.28 $\pm$ 0.27 & \cellcolor[rgb]{ .439,  .678,  .278}47.62 $\pm$ 1.27 & 29.17 $\pm$ 1.46 & \cite{rozemberczki2021multi} \\
          & {Benign} & Chameleon-filtered & 890   & 8,854 & 2,325 & 5     & 0.2361 & 0.2441 & {Accuracy} & \cellcolor[rgb]{ .439,  .678,  .278}41.46 $\pm$ 3.42 & 38.06 $\pm$ 3.98 & \cellcolor[rgb]{ .439,  .678,  .278}44.00 $\pm$ 3.10 & 35.72 $\pm$ 2.23 & \cite{platonov2022critical} \\
          &       &  arXiv-year & 169,343 & 1,166,243 & 128   & 5     & 0.2218 & 0.2778 & ROC AUC & \cellcolor[rgb]{ .439,  .678,  .278}40 $\pm$ 0.26 & 36.36 $\pm$ 0.23 & \cellcolor[rgb]{ .439,  .678,  .278}35.58 $\pm$ 0.22  & 34.11 $\pm$ 0.17 & \cite{lim2021large} \\
          &       &  amazon-ratings & 24,492 & 93,050 & 300   & 5     & 0.3804 & 0.3757 & Accuracy & \cellcolor[rgb]{ .439,  .678,  .278}50.05 $\pm$ 0.67 & 49.55 $\pm$ 0.81 & \cellcolor[rgb]{ .439,  .678,  .278}40.69 $\pm$ 0.42 & 38.60 $\pm$ 0.41 & \cite{platonov2022critical} \\
          &       & Wiki-cooc & 10,000 & 2,243,042 & 100   & 5     & 0.3435 & 0.175 & {Accuracy} & \cellcolor[rgb]{ .439,  .678,  .278}95.40 $\pm$ 0.41 & 89.38 $\pm$ 0.42 & \cellcolor[rgb]{ .439,  .678,  .278}72.38 $\pm$ 0.78 & 48.86 $\pm$ 0.37 & \cite{zhou2024opengsl} \\
\cmidrule{2-15}          &       & Squirrel-filtered & 2,223 & 46,998 & 2,089 & 5     & 0.2072 & 0.1905 & Accuracy & 37.33 $\pm$ 1.88 & \cellcolor[rgb]{ .439,  .678,  .278}38.30 $\pm$ 1.22 & \cellcolor[rgb]{ .439,  .678,  .278}37.54 $\pm$ 2.13 & 30.14 $\pm$ 1.53 & \cite{platonov2022critical} \\
          &       & Penn94 & 41,554 & 1,362,229 & 5     & 2     & 0.4704 & 0.4828 & Accuracy & \cellcolor[rgb]{ .439,  .678,  .278}82.08 $\pm$ 0.31 & 74.68 $\pm$ 0.28 & 67.06 $\pm$ 0.19 & \cellcolor[rgb]{ .439,  .678,  .278}73.72 $\pm$ 0.5 & \cite{lim2021large} \\
          & {Ambiguous} &  pokec & 1,632,803 & 30,622,564 & 65    & 2     & 0.4449 &  0.3931    & Accuracy & \cellcolor[rgb]{ .439,  .678,  .278}70.3 $\pm$ 0.1 & 62.13 $\pm$ 0.1 & 52.88 $\pm$ 0.64 & \cellcolor[rgb]{ .439,  .678,  .278}59.89 $\pm$ 0.11 & \cite{lim2021large} \\
          &       &  snap-patents & 2,923,922 & 13,975,788 & 269   & 5     & 0.073 &  0.1857     & Accuracy & \cellcolor[rgb]{ .439,  .678,  .278}35.8 $\pm$ 0.05 & 31.43 $\pm$ 0.04 & 29.65 $\pm$ 0.04 & \cellcolor[rgb]{ .439,  .678,  .278}30.59 $\pm$ 0.02 & \cite{lim2021large} \\
          &       &  twitch-gamers & 168,114 & 6,797,557 & 7     & 2     & 0.545 & 0.556 & Accuracy & \cellcolor[rgb]{ .439,  .678,  .278}62.33 $\pm$ 0.23 & 60.9 $\pm$ 0.11 & 57.9 $\pm$ 0.18 & \cellcolor[rgb]{ .439,  .678,  .278}59.45 $\pm$ 0.16 & \cite{lim2021large} \\
         \cmidrule{1-15}  &       & {Cora} & 2,708 & 5,429 & 1,433 & 7     & 0.8100 & 0.8252 & Accuracy & \cellcolor[rgb]{ .439,  .678,  .278}87.78 $\pm$ 0.96 & 76.44 $\pm$ 0.30 & \cellcolor[rgb]{ .439,  .678,  .278}85.12 $\pm$ 1.64 & 74.3 $\pm$ 1.27 & \cite{yang2016revisiting} \\
          &       & {CiteSeer} & 3,327 & 4,732 & 3,703 & 6     & 0.7355 & 0.7062 & Accuracy & \cellcolor[rgb]{ .439,  .678,  .278}81.39 $\pm$ 1.23 & 76.25 $\pm$ 0.28 & \cellcolor[rgb]{ .439,  .678,  .278}79.66 $\pm$ 0.75 & 75.51 $\pm$ 1.35 & \cite{yang2016revisiting} \\
   {Homophily } &       & {PubMed} & 19,717 & 44,338 & 500   & 3     & 0.8024 & 0.7924 & Accuracy & \cellcolor[rgb]{ .439,  .678,  .278}88.9 $\pm$ 0.32 & 86.43 $\pm$ 0.13 & \cellcolor[rgb]{ .439,  .678,  .278}86.5 $\pm$ 0.76 & 86.23 $\pm$ 0.54 & \cite{yang2016revisiting} \\
    {Graphs} &       &  minesweeper  & 10,000 & 39,402 & 7     & 2     & 0.6828 & 0.6829 & ROC AUC & \cellcolor[rgb]{ .439,  .678,  .278}72.34 $\pm$ 0.93 & 50.92 $\pm$ 1.25 & \cellcolor[rgb]{ .439,  .678,  .278}82.04 $\pm$ 0.77 & 50.59 $\pm$ 0.83 & \cite{platonov2022critical} \\
          &       &  tolokers & 11,758 & 519,000 & 10    & 2     & 0.5945 & 0.6344 & ROC AUC & \cellcolor[rgb]{ .439,  .678,  .278}77.44 $\pm$ 1.32 & 74.58 $\pm$ 0.75 & \cellcolor[rgb]{ .439,  .678,  .278}73.80 $\pm$ 1.35 & 71.89 $\pm$ 0.82 & \cite{platonov2022critical} \\
          &       &  questions & 48,921 & 153,540 & 301   & 2     & 0.8396 & 0.898 & ROC AUC & \cellcolor[rgb]{ .439,  .678,  .278}72.72 $\pm$ 1.93 & 69.97 $\pm$ 1.16 & \cellcolor[rgb]{ .439,  .678,  .278}71.06 $\pm$ 0.92 & 70.33 $\pm$ 0.96 & \cite{platonov2022critical} \\
    \bottomrule
    \bottomrule
    \end{tabular}%
    }
    \vspace{-0.2cm}
  \label{tab:categorization_homogeneous_benchmark}%
\end{table*}%

\vspace{-0.1cm}
As stated in~\cite{ma2021homophily, luan2021heterophily, luan2023graph}, heterophily does not always lead to inferior performance and homophily is not always necessary for GNNs. Therefore, empirical results are important to identify and distinguish the truly difficult and pseudo-difficult heterophily\footnote{In this paper, a dataset is considered as heterophilic if at least one of its $\text{H}_\text{edge}$ or $\text{H}_\text{node}$ value is smaller or close to $0.5$, otherwise it is homophilic.} datasets. Specifically, we conduct ablation study to evaluate the impact of graph structure on message passing (MP). For example, as shown in Section~\ref{sec:graph_aware_and_agnostic}, if we remove the MP step in GCN and the model performance decreases, this means MP with the graph structure is beneficial; otherwise, the MP with graph structure is harmful. In this way, we can reliably find out the easy and challenging heterophilic datasets. We have also compared the linear models SGC vs. MLP-1. Note that sufficient hyperparameter tuning for each baseline model is important to guarantee fair comparison. The statistics and experimental results are shown in Table~\ref{tab:categorization_homogeneous_benchmark}.  From the results, we have identified heterophilic datasets with distinct properties as follows,
\begin{itemize}
    \item \textbf{Malignant and Benign Heterophily:} There exist a subset of heterophilic datasets where the graph-aware models consistently underperform their corresponding graph-agnostic models, \eg{} \textit{Cornell, Wisconsin, Texas, Film, Deezer-Europe, genius, roman-empire, BlogCatalog, Flickr} and \textit{BGP}, which indicates that these heterophilic graph structure provides harmful information in feature aggregation step. On the other hand, there exist another class of heterophilic graphs where the graph-aware models consistently outperform graph-agnostic models, \eg{} \textit{Chameleon, Squirrel, Chameleon-filtered, arXiv-year, amazon-ratings} and \textit{Wiki-cooc}, which indicates that these heterophilic graph structures actually provide beneficial information for GNNs and do not cause any trouble to graph learning.\footnote{This is consistent with the conclusions in~\cite{ma2021homophily, luan2022revisiting, luan2023graph}} We call them \textbf{malignant and benign heterophilic datasets}, separately.
    \item \textbf{Ambiguous Heterophily:} Besides, we discover a third group of heterophily datasets, where there exists inconsistency between linear and non-linear graph-aware models compared with their coupled graph-agnostic models. For instance, on \textit{Penn94, pokec, snap-patents} and \textit{twitch-gamers}, GCN (non-linear model) outperforms MLP-2, while SGC-1 (linear model) underperforms MLP-1; on \textit{Squirrel-filtered}, GCN underperforms MLP-2 but SGC-1  outperforms MLP-1. Such inconsistency indicates that the underlying synergy between graph structure and model non-linearity can influence GNN performance together. However, no theory can explain such relationship for now. Thus, we call this group of datasets the \textbf{ambiguous heterophilic dataset.}
    \item \textbf{Remark:}  The tasks on malignant and ambiguous heterophilic dataset are considered as the truly challenging ones and benign heterophily is a pseudo challenge. Besides, some newly published heterophilic datasets are actually homophilic dataset as their homophily values are much larger than $0.5$, \eg{}\textit{minesweeper, tolokers, questions}. These homophilic and benign heterophilic datasets should not be used to evaluate GNNs on heterophily issue.
\end{itemize}

\begin{table}[htbp]
 \vspace{-0.1cm}
  \centering
  \caption{Re-evaluation and comparison of $11$ SOTA models on different categories of datasets. The result or ranking is marked by \red{red} if it is worse than the best baseline models (GCN, SGC-1, MLP-2, MLP-1); the first, second and third best average ranking is marked by \textcolor{deepgreen}{green}, \textcolor{blue}{blue}, \textcolor{orange}{orange}. OOM is short for out-of-memory.}
  \resizebox{1\hsize}{!}{
  \begin{tabular}{c|cc|cc|ccccccccccc}
    \toprule
    \toprule
    \multicolumn{3}{c|}{Methods} & \multicolumn{2}{c|}{Baseline} & \multicolumn{1}{c|}{Ego-Neighbor Sep.} & \multicolumn{3}{c|}{Negative Message Passing} & \multicolumn{1}{c|}{HP + Ada. Mixing} & \multicolumn{2}{c|}{Selective Message Passing} & \multicolumn{2}{c|}{Spectral GNN} & \multicolumn{2}{c}{Multi-hop GNN} \\
    \midrule
    \multicolumn{2}{c|}{Categories} & {Datasets} & {Best} & \multicolumn{1}{c|}{Worst} & \multicolumn{1}{c|}{H$_2$GCN} & GPRGNN & FAGCN & \multicolumn{1}{c|}{GloGNN*} & \multicolumn{1}{c|}{ACM-GCN*} & GBK-GNN & \multicolumn{1}{c|}{FSGNN} & JacobiConv & \multicolumn{1}{c|}{BernNet} & APPNP & LINKX \\
         \midrule
         & \multicolumn{1}{c|}{} & {Cornell} & 93.77 $\pm$ 3.34 & 70.98 $\pm$ 8.39 & 86.23 $\pm$ 4.71 & 91.36 $\pm$ 0.70 & 88.03 $\pm$ 5.6 & 86.32 $\pm$ 3.62 & 95.9 $\pm$ 1.83 & 80.26 $\pm$ 7.92 & 91.58 $\pm$ 4.68 & 89.74 $\pm$ 5.58 & 92.13 $\pm$ 1.64 & 87.37 $\pm$ 5.49 & 82.11 $\pm$ 4.53 \\
          & \multicolumn{1}{c|}{} & {Wisconsin} & 93.87 $\pm$ 3.33 & 70.38 $\pm$ 2.85 & 87.5 $\pm$ 1.77 & 93.75 $\pm$ 2.37 & 89.75 $\pm$ 6.37 & 89.98 $\pm$ 2.63  & 97.5 $\pm$ 1.25 & 85.10 $\pm$ 5.49 & 89.22 $\pm$ 3.19 & 88.04 $\pm$ 4.15 & \multicolumn{1}{l}{87.25 $\pm$ 3.75} & 85.29 $\pm$ 5.77 & 83.53 $\pm$ 4.74 \\
          & \multicolumn{1}{c|}{} & {Texas} & 93.77 $\pm$ 3.34 & 83.11 $\pm$ 3.2 & 85.90 $\pm$ 3.53 & 92.92 $\pm$ 0.61 & 88.85 $\pm$ 4.39 & 87.62 $\pm$ 4.89  & 96.56 $\pm$ 2 & 84.21 $\pm$ 6.12 & 90.26 $\pm$ 4.86 & 88.95 $\pm$ 5.86 & 93.12 $\pm$ 0.65 & 87.89 $\pm$ 6.02 & 84.21 $\pm$ 6.12 \\
          & \multicolumn{1}{c|}{} & {Film} & 38.58 $\pm$ 0.25 & 25.26 $\pm$ 1.18 & 38.85 $\pm$ 1.17 & 39.30 $\pm$ 0.27 & 31.59 $\pm$ 1.37 & 39.65 $\pm$ 1.03 & 41.86 $\pm$ 1.48 & 38.47 $\pm$ 1.53 & 37.65 $\pm$ 0.79 & 37.48 $\pm$ 0.76 & 41.79 $\pm$ 1.01 & 37.68 $\pm$ 0.96 & 35.64 $\pm$ 1.36 \\
          & \multicolumn{1}{c|}{Malignant} & {Deezer-Europe} & 66.55 $\pm$ 0.72 & 61.63 $\pm$ 0.25 & 67.22 $\pm$ 0.90 & 66.90 $\pm$ 0.50 & 66.86 $\pm$ 0.53 & OOM   & 67.5 $\pm$ 0.53 & OOM   & OOM   & 66.71 $\pm$ 0.6 & 67.03 $\pm$ 0.55 & 66.03 $\pm$ 0.54 & 65.76 $\pm$ 0.41 \\
          & \multicolumn{1}{c|}{} &  genius & 86.62 $\pm$ 0.08 & 82.31 $\pm$ 0.45 & 87.67 $\pm$ 0.10 &  90.05 $\pm$ 0.31 & 90.03 $\pm$ 0.20 &  90.91 $\pm$ 0.13 & 91.37 $\pm$ 0.07 & OOM   & 89.82 $\pm$ 0.03 & 86.00 $\pm$ 3.52 & 86.16 $\pm$ 0.35 & 87.61 $\pm$ 0.12 &  90.77 $\pm$ 0.27 \\
          & \multicolumn{1}{c|}{} &  roman-empire  & 66.04 $\pm$ 0.71 & 44.60 $\pm$ 0.52 & 60.11 $\pm$ 0.52 & 64.85 $\pm$ 0.27 & 65.22 $\pm$ 0.56 & 59.63 $\pm$ 0.69 & 71.89 $\pm$ 0.61 & 74.57 $\pm$ 0.47 & 79.92 $\pm$ 0.56 & 71.14 $\pm$ 0.42 & 65.56 $\pm$ 1.34 & 65.87 $\pm$ 0.53 & 56.15 $\pm$ 0.93 \\
          & \multicolumn{1}{c|}{} & BlogCatalog & 92.97 $\pm$ 0.89 & 71.07 $\pm$ 1.15 & 97.14 $\pm$ 0.50 & 97.07 $\pm$ 0.45 & 97.31 $\pm$ 0.46 & OOM   & 97.38 $\pm$ 0.41 & OOM   & 97.00 $\pm$ 0.55 & 96.84 $\pm$ 0.36 & 96.95 $\pm$ 0.52 & 96.01 $\pm$ 0.56 & 95.81 $\pm$ 0.69 \\
          & \multicolumn{1}{c|}{} & Flickr & 90.24 $\pm$ 0.96 & 60.10 $\pm$ 1.21 & 92.46 $\pm$ 1.00 & 92.60 $\pm$ 0.70 & 93.50 $\pm$ 0.81 & OOM   & 92.64 $\pm$ 0.67 & OOM   & 93.39 $\pm$ 0.99 & 92.29 $\pm$ 0.99 & 92.71 $\pm$ 0.80 & 91.43 $\pm$ 0.67 & 90.69 $\pm$ 0.73 \\
          & \multicolumn{1}{c|}{} & BGP   & 65.56 $\pm$ 0.55 & 61.74 $\pm$ 0.73 & 66.40 $\pm$ 0.73 & 65.48 $\pm$ 0.77 & 66.06 $\pm$ 0.54 & OOM   & 66.79 $\pm$ 0.81 & 66.92 $\pm$ 0.49 & 66.72 $\pm$ 0.62 & 65.51 $\pm$ 0.53 & 66.04 $\pm$ 0.66 & 65.66 $\pm$ 0.64 & 63.80 $\pm$ 0.62 \\
\cmidrule{2-16}    \multicolumn{1}{c|}{Heterophily} & \multicolumn{1}{c|}{} & {Chameleon} & 64.86 $\pm$ 1.81 & 45.01 $\pm$ 1.58 & 52.30 $\pm$ 0.48 & 67.48 $\pm$ 0.40 & 49.47 $\pm$ 2.84 & 71.98 $\pm$ 2.38 & 76.08 $\pm$ 2.13 & 50.57 $\pm$ 1.86 & 76.95 $\pm$ 1.03 & 72.11 $\pm$ 2.77 & 68.29 $\pm$ 1.58 & 48.55 $\pm$ 1.89 & 81.38 $\pm$ 1.41 \\
    \multicolumn{1}{c|}{Graphs } & \multicolumn{1}{c|}{} & {Squirrel} & 47.62 $\pm$ 1.27 & 29.17 $\pm$ 1.46 & 30.39 $\pm$ 1.22 & 49.93 $\pm$ 0.53 & 42.24 $\pm$ 1.2 & 59.56 $\pm$ 1.82 & 69.98 $\pm$ 1.53 & 34.92 $\pm$ 1.23 & 72.11 $\pm$ 2.66 & 55.86 $\pm$ 1.46 & 51.35 $\pm$ 0.73 & 34.08 $\pm$ 1.21 & 77.44 $\pm$ 1.69 \\
          & \multicolumn{1}{c|}{Benign} & Chameleon-filtered & 44.00 $\pm$ 3.10 & 35.72 $\pm$ 2.23 & 42.90 $\pm$ 3.91 & 41.95 $\pm$ 3.68 & 42.87 $\pm$ 5.01 & OOM   & 42.73 $\pm$ 3.59 & 36.20 $\pm$ 4.37 & 40.96 $\pm$ 2.73 & 41.42 $\pm$ 2.67 & 40.90 $\pm$ 4.06 & 37.50 $\pm$ 3.69 & 42.34 $\pm$ 4.13 \\
          & \multicolumn{1}{c|}{} &  arXiv-year & 40 $\pm$ 0.26 & 34.11 $\pm$ 0.17 &  49.09 $\pm$ 0.10 &  45.07 $\pm$ 0.21 & 40.12 $\pm$ 0.44 &  54.79 $\pm$ 0.25 & 52.49 $\pm$ 0.23 & OOM   & 50.62 $\pm$ 0.18 & 38.07 $\pm$ 0.21 & \multicolumn{1}{l}{35.62 $\pm$ 0.19} & 35.23 $\pm$ 0.16 &  56.00 $\pm$ 1.34 \\
          & \multicolumn{1}{c|}{} &  amazon-ratings & 50.05 $\pm$ 0.67 & 38.60 $\pm$ 0.41 & \textcolor[rgb]{ 1,  0,  0}  {36.47 $\pm$ 0.23} &  44.88 $\pm$ 0.34 &  44.12 $\pm$ 0.30 & \textcolor[rgb]{ 1,  0,  0}{36.89 $\pm$ 0.14}  & 52.49 $\pm$ 0.24 &  45.98 $\pm$ 0.71 &  52.74 $\pm$ 0.83 &  43.55 $\pm$ 0.48  & 44.64 $\pm$ 0.56 & 46.02 $\pm$ 0.73 & 52.66 $\pm$ 0.64 \\
          & \multicolumn{1}{c|}{} & Wiki-cooc & 95.40 $\pm$ 0.41 & 48.86 $\pm$ 0.37 & 98.75 $\pm$ 0.15 & 92.58 $\pm$ 1.18 & 89.50 $\pm$ 0.92 & OOM   & 99.32 $\pm$ 0.23 & OOM   & 98.96 $\pm$ 0.15 & 90.22 $\pm$ 0.6 & 94.76 $\pm$ 0.31 & 88.96 $\pm$ 0.49 & 98.24 $\pm$ 0.37 \\
\cmidrule{2-16}          & \multicolumn{1}{c|}{} & Squirrel-filtered & 38.30 $\pm$ 1.22 & 30.14 $\pm$ 1.53 & 42.77 $\pm$ 1.61 & 38.05 $\pm$ 1.44 & 42.37 $\pm$ 1.77 & OOM   & 42.35 $\pm$ 1.97 & 35.07 $\pm$ 1.26 & 37.56 $\pm$ 1.12 & 42.71 $\pm$ 1.75 & 41.18 $\pm$ 1.77 & 35.12 $\pm$ 1.12 & 40.10 $\pm$ 2.21 \\
          & \multicolumn{1}{c|}{} & Penn94 & 82.08 $\pm$ 0.31 & 67.06 $\pm$ 0.19 & 81.31 $\pm$ 0.60 & 81.38 $\pm$ 0.16 & 79.87 $\pm$ 0.82 & 85.74 $\pm$ 0.42 & 86.08 $\pm$ 0.43 & OOM   & OOM   & 83.80 $\pm$ 0.33 & 82.88 $\pm$ 0.52 & 75.57 $\pm$ 0.26 & 84.71 $\pm$ 0.52 \\
          & \multicolumn{1}{c|}{Ambiguous} &  pokec & 70.3 $\pm$ 0.1 & 52.88 $\pm$ 0.64 &  OOM  &  78.83 $\pm$ 0.05 & OOM   &  83.05 $\pm$ 0.07 & 81.07 $\pm$ 0.165 & OOM   & OOM   & 75.85 $\pm$ 0.06 & OOM   & 61.82 $\pm$ 0.19 &  82.04 $\pm$ 0.07 \\
          & \multicolumn{1}{c|}{} &  snap-patents & 35.8 $\pm$ 0.05 & 29.65 $\pm$ 0.04 &  OOM  &  40.19 $\pm$ 0.03 & OOM   & 62.09 $\pm$ 0.27 & 54.79 $\pm$ 0.616 & OOM   & OOM   & 40.87 $\pm$ 0.04 & OOM   & 32.47 $\pm$ 0.11 &  61.95 $\pm$ 0.12 \\
          & \multicolumn{1}{c|}{} &  twitch-gamers & 62.33 $\pm$ 0.23 & 57.9 $\pm$ 0.18 &  OOM  &  61.89 $\pm$ 0.29 & OOM   &  66.34 $\pm$ 0.29 & 66.24 $\pm$ 0.24 & OOM   & 61.71 $\pm$ 0.24 & 61.98 $\pm$ 0.06 & 60.08 $\pm$ 0.29 & 60.36 $\pm$ 0.18 &  66.06 $\pm$ 0.19 \\
             \midrule
          & \multicolumn{1}{c|}{} & {Cora} & 87.78 $\pm$ 0.96 & 74.3 $\pm$ 1.27 & 87.52 $\pm$ 0.61 & 79.51 $\pm$ 0.36 & 88.85 $\pm$ 1.36 & 87.67 $\pm$ 1.16 & 89.75 $\pm$ 1.16 & 87.09 $\pm$ 1.52 & 87.51 $\pm$ 1.21 & 89.61 $\pm$ 0.96 & 88.52 $\pm$ 0.95 & 88.29 $\pm$ 1.24 & 82.62 $\pm$ 1.44 \\
          & \multicolumn{1}{c|}{} & {CiteSeer} & 81.39 $\pm$ 1.23 & 75.51 $\pm$ 1.35 & 79.97 $\pm$ 0.69 & \textcolor[rgb]{ 1,  0,  0}{67.63 $\pm$ 0.38} & 82.37 $\pm$ 1.46 & 78.91 $\pm$ 1.75  & 81.87 $\pm$ 1.38 & 76.62 $\pm$ 0.84 & 76.59 $\pm$ 1.45 & 77.60 $\pm$ 1.12 & 80.09 $\pm$ 0.79 & \textcolor[rgb]{ 1,  0,  0}{74.88 $\pm$ 1.27} & \textcolor[rgb]{ 1,  0,  0}{69.92 $\pm$ 1.36} \\
    \multicolumn{1}{c|}{Homophily} & \multicolumn{1}{r|}{} & {PubMed} & 88.9 $\pm$ 0.32 & 86.23 $\pm$ 0.54 & 87.78 $\pm$ 0.28 & \textcolor[rgb]{ 1,  0,  0}{85.07 $\pm$ 0.09} & 89.98 $\pm$ 0.54 & 90.32 $\pm$ 0.54 & 90.96 $\pm$ 0.62 & 88.88 $\pm$ 0.44 & 90.11 $\pm$ 0.43 & 89.99 $\pm$ 0.39 & 88.48 $\pm$ 0.41 & 90.02 $\pm$ 0.43 & 88.12 $\pm$ 0.47 \\
    \multicolumn{1}{c|}{Graphs} & \multicolumn{1}{c|}{} &  minesweeper  & 82.04 $\pm$ 0.77 & 50.59 $\pm$ 0.83 &  89.71 $\pm$ 0.31  &  86.24 $\pm$ 0.61  &  88.17 $\pm$ 0.73  & 51.08 $\pm$ 1.23  & 84.71 $\pm$ 0.85 &  90.85 $\pm$ 0.58 &  90.08 $\pm$ 0.70 & 89.66 $\pm$ 0.40 & 77.99 $\pm$ 0.95 & 69.62 $\pm$ 2.11 & 56.78 $\pm$ 2.47 \\
          & \multicolumn{1}{c|}{} &  tolokers & 77.44 $\pm$ 1.32 & 71.89 $\pm$ 0.82 & 73.35 $\pm$ 1.01  & 72.94 $\pm$ 0.97 & 77.75 $\pm$ 1.05 & 73.39 $\pm$ 1.17  & 74.95 $\pm$ 1.16 &  81.01 $\pm$ 0.67  &  82.76 $\pm$ 0.61 & \textcolor[rgb]{ 1,  0,  0}{68.66 $\pm$ 0.65}  & 77.00 $\pm$ 0.65 & 76.98 $\pm$ 1.03 & 81.15 $\pm$ 1.23 \\
          & \multicolumn{1}{c|}{} &  questions & 72.72 $\pm$ 1.93 & 69.97 $\pm$ 1.16 & \textcolor[rgb]{ 1,  0,  0} {63.59 $\pm$ 1.46} & \textcolor[rgb]{ 1,  0,  0}{55.48 $\pm$ 0.91} &  77.24 $\pm$ 1.26 & \textcolor[rgb]{ 1,  0,  0}{65.74 $\pm$ 1.19} & \textcolor[rgb]{ 1,  0,  0}{62.91 $\pm$ 2.10} & 74.47 $\pm$ 0.86 &  78.86 $\pm$ 0.92 & 73.88 $\pm$ 1.16 & 70.43 $\pm$ 1.38 & \textcolor[rgb]{ 1,  0,  0}{64.77 $\pm$ 1.32} & 71.96 $\pm$ 2.07 \\
    \midrule
          & \multicolumn{2}{c|}{Overall} & 5.85  & 11.48 & \textcolor[rgb]{ 1,  0,  0}{6.92} & \textcolor[rgb]{ 1,  0,  0}{7.07} & 5.71  & \cellcolor[rgb]{ .929,  .49,  .192} \textbf{5.55} & \cellcolor[rgb]{ .439,  .678,  .278} \textbf{2.85} & \textcolor[rgb]{ 1,  0,  0}{7.35} & \cellcolor[rgb]{ .357,  .608,  .835} \textbf{4.39} & \textcolor[rgb]{ 1,  0,  0}{6.33} & \textcolor[rgb]{ 1,  0,  0}{6.20} & \textcolor[rgb]{ 1,  0,  0}{8.41} & \textcolor[rgb]{ 1,  0,  0}{6.67} \\
    \multicolumn{1}{c|}{Average} & \multicolumn{2}{c|}{Malignant Heterophily} & 6.10  & 12.10 & \textcolor[rgb]{ 1,  0,  0}{6.50} & 5.30  & 5.60  & \textcolor[rgb]{ 1,  0,  0}{6.33} & \cellcolor[rgb]{ .439,  .678,  .278} \textbf{1.60} & \textcolor[rgb]{ 1,  0,  0}{7.33} & \cellcolor[rgb]{ .357,  .608,  .835} \textbf{4.56} & \textcolor[rgb]{ 1,  0,  0}{7.30} & \cellcolor[rgb]{ .929,  .49,  .192} \textbf{5.20} & \textcolor[rgb]{ 1,  0,  0}{7.90} & \textcolor[rgb]{ 1,  0,  0}{9.80} \\
    \multicolumn{1}{c|}{Ranking} & \multicolumn{2}{c|}{Benign Heterophily} & 5.67  & 12.00 & \textcolor[rgb]{ 1,  0,  0}{7.33} & \textcolor[rgb]{ 1,  0,  0}{6.67} & \textcolor[rgb]{ 1,  0,  0}{8.00} & \textcolor[rgb]{ 1,  0,  0}{5.75} & \cellcolor[rgb]{ .357,  .608,  .835} \textbf{2.83} & \textcolor[rgb]{ 1,  0,  0}{9.25} & \cellcolor[rgb]{ .929,  .49,  .192} \textbf{3.17} & \textcolor[rgb]{ 1,  0,  0}{7.17} & \textcolor[rgb]{ 1,  0,  0}{7.50} & \textcolor[rgb]{ 1,  0,  0}{9.83} & \cellcolor[rgb]{ .439,  .678,  .278} \textbf{2.33} \\
          & \multicolumn{2}{c|}{Ambiguous Heterophily} & 5.80  & 9.80  & 4.50  & \textcolor[rgb]{ 1,  0,  0}{6.00} & \textcolor[rgb]{ 1,  0,  0}{6.00} & \cellcolor[rgb]{ .439,  .678,  .278} \textbf{1.25} & \cellcolor[rgb]{ .357,  .608,  .835} \textbf{2.60} & \textcolor[rgb]{ 1,  0,  0}{11.00} & \textcolor[rgb]{ 1,  0,  0}{8.00} & 4.00  & \textcolor[rgb]{ 1,  0,  0}{6.33} & \textcolor[rgb]{ 1,  0,  0}{8.40} & \cellcolor[rgb]{ .929,  .49,  .192} \textbf{3.20} \\
          & \multicolumn{2}{c|}{Homophily} & 5.67  & 11.33 & \textcolor[rgb]{ 1,  0,  0}{8.00} & \textcolor[rgb]{ 1,  0,  0}{11.33} & \cellcolor[rgb]{ .439,  .678,  .278} \textbf{3.50} & \textcolor[rgb]{ 1,  0,  0}{7.50} & \cellcolor[rgb]{ .929,  .49,  .192} \textbf{5.17} & 5.50  & \cellcolor[rgb]{ .357,  .608,  .835} \textbf{4.17} & \textcolor[rgb]{ 1,  0,  0}{5.83} & \textcolor[rgb]{ 1,  0,  0}{6.50} & \textcolor[rgb]{ 1,  0,  0}{7.83} & \textcolor[rgb]{ 1,  0,  0}{8.67} \\
    \bottomrule
    \bottomrule
    \end{tabular}%
    }
    \vspace{-0.2cm}
  \label{tab:sota_models_reassess}%
\end{table}%

\vspace{-0.1cm}
\vspace{-0.2cm}
\subsection{Reassessment of State-of-the-arts Models}
\vspace{-0.1cm}
\label{sec:reassessment_sota}
Based on the new categorization of heterophilic datasets, our next question is: \textbf{how do current SOTA models perform on different groups of datasets, and are they really effective on heterophily?} In this subsection, we reassess $11$ popular SOTA GNNs, covering six most popular methods for heterophily, with fine-tuned hyperparameters.\footnote{See Appendix~\ref{appendix:hyperparameter_searching_range} for the hyperparameter searching range.} The methods and models include: ego-neighbor separation (H$_2$GCN~\cite{zhu2020beyond}), negative message passing (GPRGNN~\cite{chien2021adaptive}, FAGCN~\cite{bo2021beyond}, GloGNN*~\cite{li2022finding} \footnote{GloGNN* indicates the best results of the variants of GloGNN.}), high-pass filter with adaptive channel mixing (ACM-GCN*~\cite{luan2022revisiting}\footnote{ACM-GCN has lots of variants, we report the best results of them as ACM-GCN*.}), selective message passing (GBK-GNN~\cite{du2022gbk}, FSGNN~\cite{maurya2022simplifying}), spectral GNN (JacobConv~\cite{wang2022howpowerful}, BernNet~\cite{he2021bernnet}), multi-hop GNN (APPNP~\cite{gasteiger2018predict}, LINKX~\cite{lim2021new}). A  model is considered effective for heterophily if it: 1) outperforms the best baseline model on malignant and ambiguous heterophily graphs, and 2) performs at least as good as the best baseline model in other categories. According to Table~\ref{tab:sota_models_reassess}, we find,
\begin{itemize}
    \item \textbf{Effectiveness of Heterophily-specific Methods:} In most tasks, the majority of SOTA GNNs do not significantly outperform the best baseline models. ACM-GCN*, FSGNN and GloGNN* are the only three models that have better overall average ranking  than the best baseline models. This means only high-pass filter with adaptive channel mixing, selective message passing and negative message passing methods are verified to have the potential to be effective for heterophily. 
    \item \textbf{Detailed Analysis:} Besides the aforementioned three methods, others are only partially effective on one of the heterophily categories, but have bad overall rankings. For example, H$_2$GCN, JacobConv and LINKX perform well on ambiguous heterophily, BernNet is good at malignant heterophily. However, they still underperform the best baselines in general. This demonstrates the necessity of comprehensive evaluation for each category. Otherwise, the conclusions about effectiveness can be questionable and unreliable. Particularly, for multi-hop GNNs, LINKX (hop=$2$) works well for ambiguous and benign heterophily, but APPNP (hop=$10$) struggles on them. This echoes the over-globalizing phenomenon~\cite{xing2024less, yadati2025localformer}, where the model focuses too much on distant nodes, but the long-range dependencies are found not to be quite informative for heterophilic graphs~\cite{xing2024less, yadati2025localformer}. In other words, we need to be fairly selective for the diffusion scopes of our model~\cite{lu2024flexible}.

    \item \textbf{Imbalanced Performance:} Some GNNs sacrifice their abilities on easy (homophily or benign heterophily) graphs to gain performance enhancement on difficult (malignant or ambiguous heterophily) graphs, \eg{} every model with negative message passing yield subpar results on benign heterophily, H$_2$GCN, GPRGNN,  GloGNN*, spectral GNNs and multi-hop GNNs show inferior results on homophily datasets. Such imbalanced performance implies that they are not universally effective methods. Surprisingly, GBK-GNN only performs well on homophily datasets, and such results is caused by the unrigorous and insufficient evaluation in the original paper. 
    \item \textbf{Scalability Issue:} Some of the tested GNNs suffer from severe out-of-memory (OOM) problem, \eg{} GloGNN, GBK-GNN and FSGNN, which indicates that some heterophily-specific methods encounter scalability issue. 
\end{itemize}

\vspace{-0.3cm}
\section{Quantitative Evaluation of Homophily Metrics on Synthetic Graphs}
\vspace{-0.2cm}
\label{sec:synthetic_comparison}
Homophily metrics are proposed to help people recognize the challenging heterophilic datasets without training models~\cite{luan2023graph}, and people usually evaluate the metrics by synthetic graphs. In Section~\ref{sec:synthetic_graph_generation}, we summarize three most widely used synthetic graph generation methods; in Section~\ref{sec:homophily_metric_evaluation_comparison}, we unify the evaluation methods, compare $11$ popular homophily metrics on synthetic graphs and illustrate the challenges of the observation-based evaluation approach; to compare the metrics strictly, in Section~\ref{sec:quantitative_benchmark_metrics}, we conduct quantitative comparisons among metrics based on Pearson correlation and Fréchet distance and provide detailed analysis.
\vspace{-0.2cm}
\subsection{Generation Methods for Synthetic Graphs}
\vspace{-0.1cm}
\label{sec:synthetic_graph_generation}

\begin{figure}[htbp!]
\vspace{-0.1cm}
    \centering
     {
     \subfloat[RG: Model  Curves]{
     \captionsetup{justification = centering}
     \includegraphics[width=0.33\textwidth]{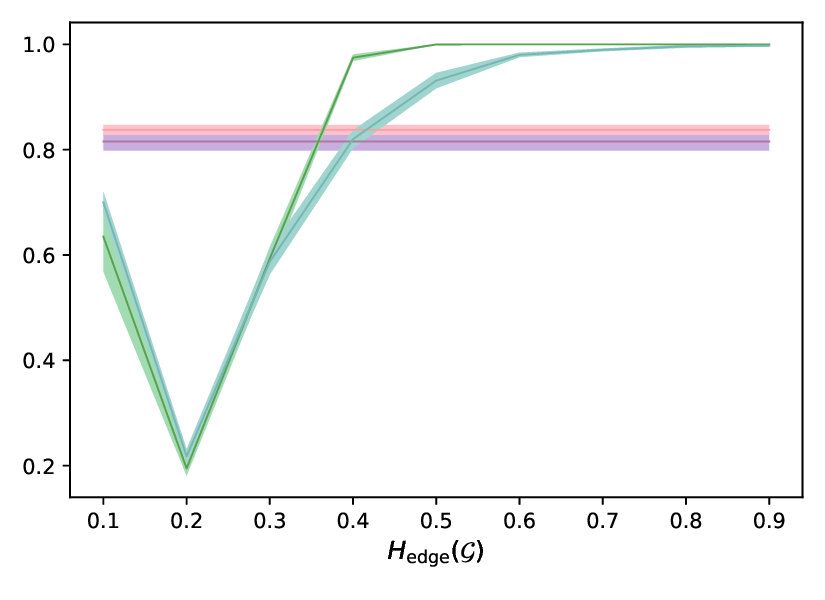}}
    \subfloat[PA: Model  Curves]{
     \captionsetup{justification = centering}
     \includegraphics[width=0.33\textwidth]{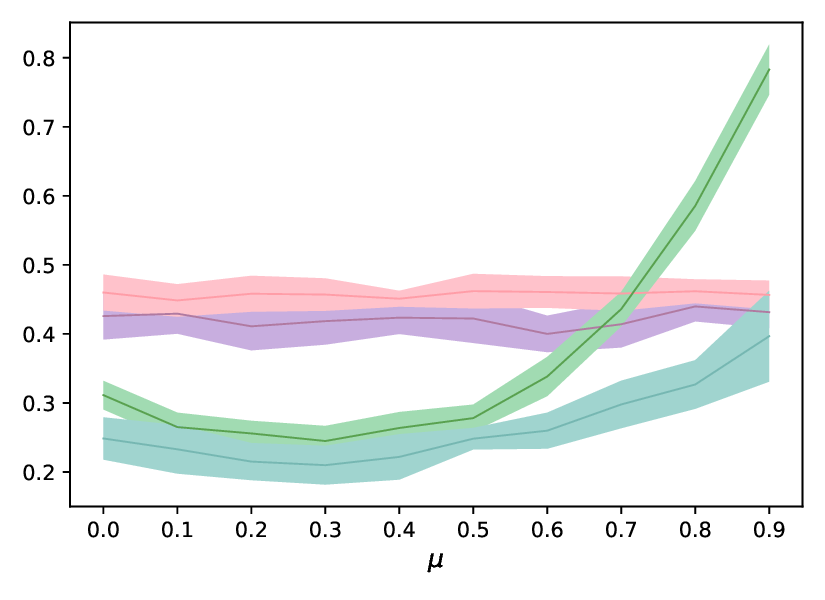}
     }
     \subfloat[GenCat: Model  Curves]{
     \captionsetup{justification = centering}
     \includegraphics[width=0.33\textwidth]{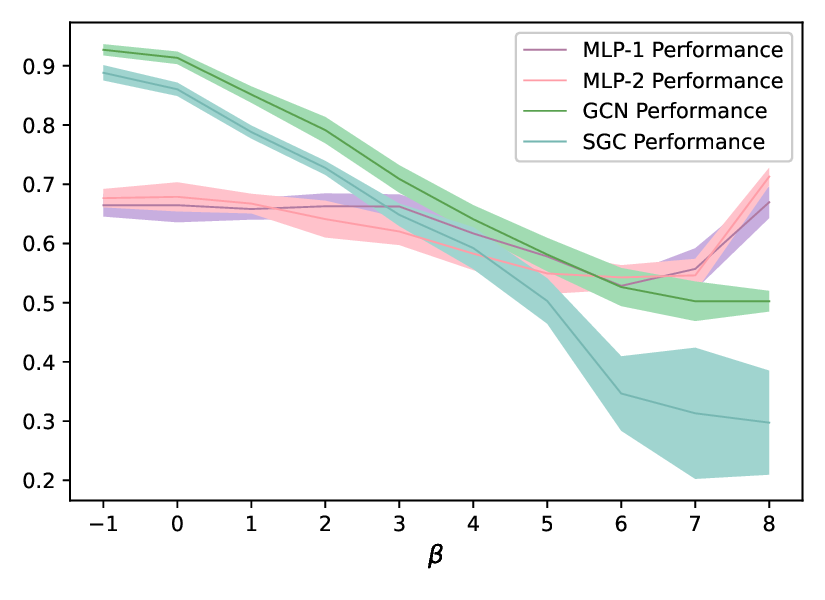}
     }\\
     \vspace{-0.5cm}
     \subfloat[RG: Metric Curves]{
     \captionsetup{justification = centering}
     \includegraphics[width=0.33\textwidth]{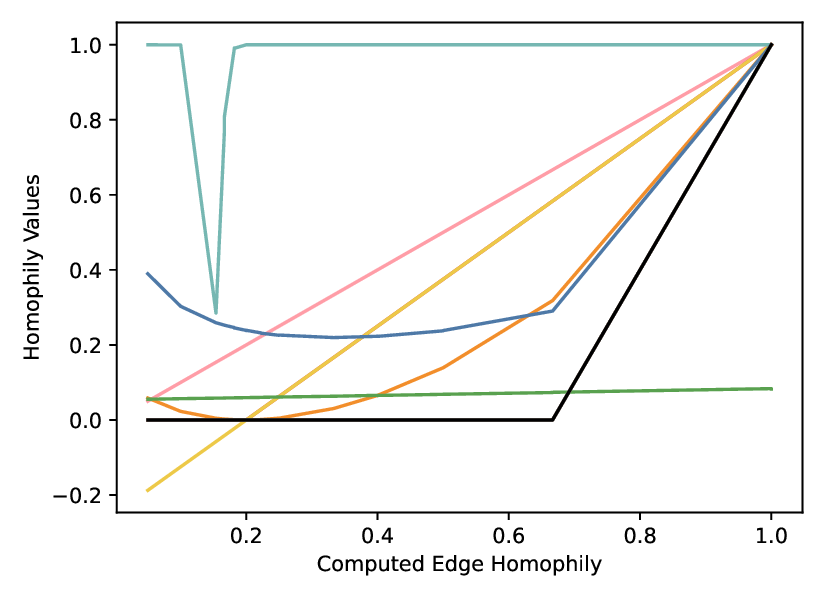}
     }
     \subfloat[PA: Metric Curves]{
     \captionsetup{justification = centering}
     \includegraphics[width=0.33\textwidth]{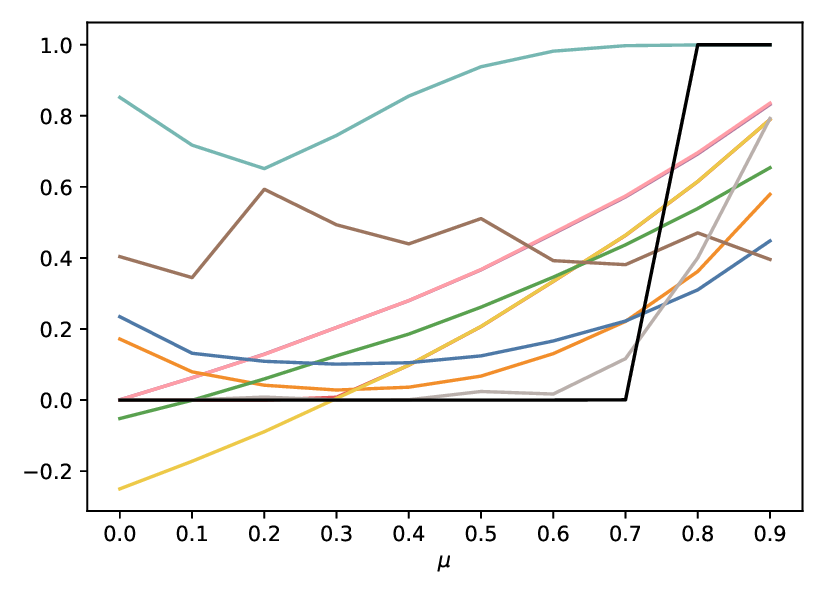}
     }
     \subfloat[GenCat: Metric Curves]{
     \captionsetup{justification = centering}
     \includegraphics[width=0.33\textwidth]{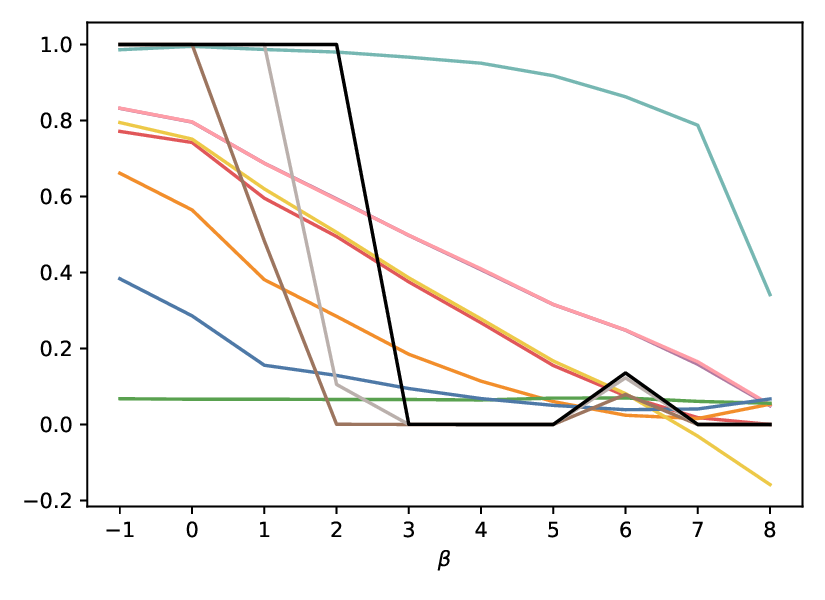}
     }
     \subfloat{
    \includegraphics[width=0.1\textwidth]{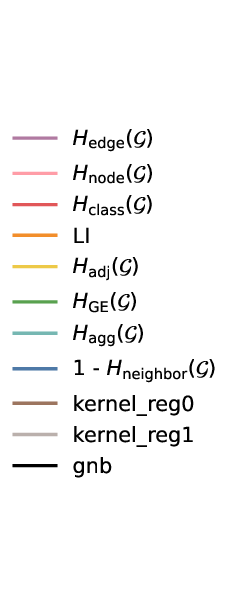}
     }
     }

     \caption{Comparison of metrics on synthetic graphs with different generation methods. Note that $\text{H}_{\text{node}}$ overlaps with $\text{H}_{\text{edge}}$ in Figure (e) and (f). In Figure (e), $\text{H}_{\text{class}}(\mathcal{G})$ overlaps with $\text{H}_{\text{adj}}(\mathcal{G})$. $\text{KR}_\text{L}$, $\text{KR}_\text{NL}$ and GNB overlaps in Figure (d).}
     \vspace{-0.2cm}
     \label{fig:comparison_metrics}
\end{figure}

There are mainly three ways to generate synthetic graphs for homophily metric evaluation.
\vspace{-0.1cm}
\paragraph{Regular Graph (RG)} Luan \etal~\cite{luan2022revisiting} proposed to generate regular graphs as follows: 1) $10$ graphs are generated for each of the $28$ edge homophily levels, from $0.005$ to $0.95$, with a total of $280$ graphs; 2) Every generated graph has five classes, with $400$ nodes in each class. For nodes in each class, $800$ random intra-class edges and [$\frac{800}{\text{H}_\text{edge}(\mathcal{G})} -800$] inter-class edges are uniformly generated ; 3) The features of nodes in each class are sampled from node features in the corresponding class of the base datasets, \eg{} Figure~\ref{fig:comparison_metrics} (a)(d) are based on the node features from \textit{Cora}. 
\vspace{-0.1cm}
\paragraph{Preferential Attachment (PA) ~\cite{barabasi1999emergence}} Karimi \etal~\cite{karimi2018homophily} incorporate homophily as an additional parameter to Preferential Attachment (PA) model and Abu-El-Haija \etal~\cite{abu2019mixhop} extend it to multi-class settings, which is widely used in graph machine learning community. The process are as follows.

Suppose graph $\mathcal{G}$ has a total number of $N$ nodes, $C$ classes, and a homophily coefficient $\mu$, the generation begins by dividing the $N$ nodes into $C$ equal-sized classes. Then, $\mathcal{G}$ (initially empty) is updated iteratively. At each step, a new node $v_i$ is added, and its class $y_i$ is randomly assigned from the set $\{1, \ldots, C\}$. Whenever a new node $v_i$ is added to $\mathcal{G}$, a connection between $v_i$ and an existing node $v_j$ in $\mathcal{G}$ is established based on the probability $\bar{p}_{ij}$, which is calculated as follows,
\begin{equation}
\label{eq:puv}
p_{ij} = \begin{cases}
d_j \times \mu, & \mbox{if $y_i = y_j$} \\
d_j \times (1-\mu) \times w_{d(y_i, y_j)}, & \mbox{otherwise}
\end{cases}
, \text{ and }\;
\bar{p}_{ij} = \frac{p_{ij}}{\sum_{k: v_k \in \mathcal{N}(v_i)} p_{ik}}
\end{equation}
where $y_i$ and $y_j$ are class labels of node $i$ and $j$ respectively, and $w_{d(y_i, y_j)}$\footnote{The code for calculating $w_{d(y_i, y_j)}$ is not open-sourced and we obtain the code from the authors of~\cite{abu2019mixhop}.} denotes the ``cost'' of connecting nodes from two distinct classes with a class distance of $d(y_i, y_j)$. \footnote{The distance between two classes simply implies the shortest distance between the two classes on a circle where classes are numbered from 1 to $C$.  For instance, if $C = 6$, $y_i = 1$ and $y_j = 5$, then the distance between $y_i$ and $y_j$ is $2$.} The weight exponentially decreases as the distance increases and is normalized such that $\sum_d w_d = 1$.  For a larger $\mu$, the chance of connecting with a node with the same label increases. 
Lastly, the features of each node in the output graph are sampled from overlapping 2D Gaussian distributions. Each class has its own distribution defined separately.

\vspace{-0.1cm}
\paragraph{GenCat} GenCat~\cite{maekawa2022beyond,maekawa2023gencat} generates synthetic graphs based on a real-world graph and a hyper-parameter $\beta$ controlling the homophily/heterophily property of the generated graph.
According to base graph and $\beta$,
class preference mean ${M}^{(\beta)} \in \Rbb^{C\times C}$,
class preference deviation ${D}^{(\beta)} \in \Rbb^{C\times C}$,
class size distribution and
attribute-class correlation $H \in \Rbb^{F\times C}$ are calculated,
which are then used to create three latent factors: node-class membership proportions  ${U} \in [0, 1]^{N\times C}$, node-class connection proportions ${U}' \in [0, 1]^{N\times C}$, 
and attribute-class proportions ${V} \in [0, 1]^{F\times C}$, where $C, F$ and $N$ are the numbers of classes, features and nodes of the base graph, respectively. Finally, the synthetic graph is generated using these latent factors.

The class preference mean between class $c_1$ and class $c_2$ is initially calculated as:
$$
{M}_{c_1, c_2} = \frac{1}{|\Omega_{c_1}|} \sum_{i \in \Omega_{c_1}} \left ( \sum_{j \in \Omega_{c_2}} {A}_{ij} / \sum_j {A}_{ij} \right ),
$$
where $\Omega_{c_k} = \{v | {Z}_{v, k} = 1 \}$ is the set of nodes in class $c_k$. Then, ${M}_{c_1, c_2}$ is adjusted by $\beta$ as follows,
$$
{M}_{c_1, c_2}^{(\beta)} = \begin{cases}
\max({M}_{c_1, c_2} - 0.1 * \beta, 0) & (c_1 = c_2)\\
{M}_{c_1, c_2} + 0.1 * \beta/(C-1) & (c_1 \neq c_2)
\end{cases}.
$$
For a larger $\beta$, fewer edges would be generated later between nodes within the same class, thus corresponding to a more heterophilic graph. The range of $\beta$ is $\{\left \lfloor 10 {M}_{\text{avg}} \right \rfloor -9, \left \lfloor 10 {M}_{\text{avg}} \right \rfloor -8, \ldots, \left \lfloor 10 {M}_{\text{avg}} \right \rfloor \}$. The average of intra-class connections is calculated as ${M}_{\text{avg}} = \frac{1}{C} \sum_{c_i} {M}_{c_i, c_i}$.

\vspace{-0.2cm}
\subsection{Evaluation of Metrics and Observation-Based Comparison}
\vspace{-0.1cm}
\label{sec:homophily_metric_evaluation_comparison}

\paragraph{Evaluation Process}
It includes the following steps: 1) generate synthetic graphs with different homophily-related hyperparameters, \eg{} $\text{H}_\text{edge}$ for regular graphs, $\mu$ for PA model and $\beta$ for GenCat; 2) split nodes randomly into 60\%/20\%/20\% for train/validation/test; 3) each baseline model (GCN, SGC-1, MLP-2 and MLP-1) is trained on every synthetic graph with the same hyperparameter searching range as~\cite{luan2022revisiting}, the mean test accuracy and standard deviation of $10$ runs are recorded; 4) calculate the corresponding metric values for each synthetic graph; 5) plot the metric curves and the model performance curves \wrt{} the homophily-related hyperparameters, observe their relations.
\vspace{-0.1cm}

The model performance curves and the metric curves are shown in Figure~\ref{fig:comparison_metrics} (a)(b)(c)  and Figure~\ref{fig:comparison_metrics} (d)(e)(f). A metric is considered superior to another if the shape of its curve aligns more closely with GNN performance curves. 
\paragraph{Challenges of Observation-based Comparison} It is hard to justify which metric is superior based merely on observation. For example, in PA generated graphs, GNN performance exhibits a bit of bouncing back in low-homophily area. However, $\text{H}_\text{agg}, \text{H}_\text{neighbor}$ and $\text{LI}$ all exhibit such shape and it is difficult to determine which is better. Therefore, quantitative evaluation is required for strict and reliable comparison and we will provide the first quantitative evaluation on homophily metrics in the next subsection.
\vspace{-0.3cm}
\subsection{Quantitative Evaluation for Homophily Metrics}
\vspace{-0.1cm}
\label{sec:quantitative_benchmark_metrics}
\begin{table}[htbp]
  \centering
   \vspace{-0.2cm}
  \caption{Quantitative comparison of homophily metrics on synthetic graphs. Cells marked by \textcolor{deepgreen}{green} and \red{red} are the best and worst ranked metrics.}
      \resizebox{1\hsize}{!}{
 \begin{tabular}{c|cc|cc|cc|cc|cc|cc|cccccc}
    \toprule
    \toprule
    \multirow{3}[6]{*}{Metrics/Graphs} & \multicolumn{4}{c|}{RG}       & \multicolumn{4}{c|}{PA}       & \multicolumn{4}{c|}{GenCat}   & \multicolumn{6}{c}{Average Ranking} \\
\cmidrule{2-19}          & \multicolumn{2}{c|}{Pearson} & \multicolumn{2}{c|}{Fréchet} & \multicolumn{2}{c|}{Pearson} & \multicolumn{2}{c|}{Fréchet} & \multicolumn{2}{c|}{Pearson} & \multicolumn{2}{c|}{Fréchet} & \multicolumn{2}{c|}{RG} & \multicolumn{2}{c|}{PA} & \multicolumn{2}{c}{GenCat} \\
\cmidrule{2-19}          & GCN   & SGC   & GCN   & SGC   & GCN   & SGC   & GCN   & SGC   & GCN   & SGC   & GCN   & SGC   & Pearson & \multicolumn{1}{c|}{Fréchet} & Pearson & \multicolumn{1}{c|}{Fréchet} & Pearson & Fréchet \\
    \midrule
    $\text{H}_\text{edge}$ & 0.75  & 0.78  & 0.53  & 0.60  & 0.86  & 0.87  & 0.19  & 0.17  & 0.99  & 0.99  & 0.27  & 0.21  & 8.00  & 4.00  & 5.00  & 3.00  & \cellcolor[rgb]{ .439,  .678,  .278} \textbf{1.50} & 5.00 \\
  $\text{H}_\text{node}$ & 0.76  & 0.79  & 0.51  & 0.58  & 0.86  & 0.87  & 0.19  & 0.17  & 0.99  & 0.99  & 0.27  & 0.21  & 6.00  & 2.00  & 5.00  & 4.00  & \cellcolor[rgb]{ .439,  .678,  .278} \textbf{1.50} & 4.00 \\
   $\text{H}_\text{class}$ & 0.62  & 0.69  & 0.63  & 0.69  & 0.94  & 0.95  & 0.07  & 0.15  & 1.00  & 0.98  & 0.15  & 0.21  & 9.50  & 6.50  & 3.00  & \cellcolor[rgb]{ .439,  .678,  .278} \textbf{1.00} & 2.50  & 2.50 \\
   $\text{H}_\text{adj}$ & 0.75  & 0.78  & 0.76  & 0.83  & 0.86  & 0.87  & 0.19  & 0.17  & 0.99  & 0.99  & 0.27  & 0.21  & 7.00  & \cellcolor[rgb]{ 1,  0,  0} 11.00 & 5.00  & 2.00  & \cellcolor[rgb]{ .439,  .678,  .278} \textbf{1.50} & 4.50 \\
   $\text{H}_\text{GE}$ & 0.26  & 0.30  & 0.64  & 0.65  & 0.87  & 0.88  & 0.94  & 0.92  & 0.42  & 0.46  & 0.26  & 0.21  & \cellcolor[rgb]{ 1,  0,  0} 11.00 & 7.50  & 4.00  & \cellcolor[rgb]{ 1,  0,  0} 11.00 & \cellcolor[rgb]{ 1,  0,  0} 11.00 & 3.00 \\
   $\text{H}_{\text{agg}}$ & 0.82  & 0.83  & 0.37  & 0.30  & 0.60  & 0.64  & 0.51  & 0.46  & 0.84  & 0.90  & 0.42  & 0.40  & 5.00  & \cellcolor[rgb]{ .439,  .678,  .278} \textbf{1.00} & 10.00 & 7.00  & 7.00  & 9.50 \\
   $\text{LI}$ & 0.62  & 0.69  & 0.61  & 0.68  & 0.99  & 0.98  & 0.28  & 0.34  & 0.95  & 0.91  & 0.14  & 0.12  & 9.50  & 5.50  & \cellcolor[rgb]{ .439,  .678,  .278} \textbf{1.00} & 5.00  & 5.00  & \cellcolor[rgb]{ .439,  .678,  .278} \textbf{1.00} \\
   $\text{H}_{\text{neighbor}}$ & 0.88  & 0.87  & 0.51  & 0.58  & 0.97  & 0.96  & 0.50  & 0.50  & 0.87  & 0.82  & 0.26  & 0.18  & 2.50  & 3.00  & 2.00  & 7.50  & 7.00  & 3.00 \\
   $\text{GNB}$ & 0.91  & 0.85  & 0.63  & 0.70  & 0.80  & 0.78  & 0.48  & 0.50  & 0.87  & 0.81  & 0.38  & 0.48  & \cellcolor[rgb]{ .439,  .678,  .278} \textbf{1.50} & 7.50  & 9.00  & 6.50  & 8.00  & 9.50 \\
   $\text{KR}_\text{L}$ & 0.91  & 0.85  & 0.63  & 0.70  & -0.23 & -0.22 & 0.80  & 0.77  & 0.69  & 0.77  & 0.88  & 0.79  & \cellcolor[rgb]{ .439,  .678,  .278} \textbf{1.50} & 7.50  & \cellcolor[rgb]{ 1,  0,  0} 11.00 & 10.00 & 10.00 & \cellcolor[rgb]{ 1,  0,  0} 11.00 \\
   $\text{KR}_\text{NL}$ & 0.91  & 0.85  & 0.63  & 0.70  & 0.83  & 0.84  & 0.67  & 0.60  & 0.74  & 0.82  & 0.36  & 0.27  & \cellcolor[rgb]{ .439,  .678,  .278} \textbf{1.50} & 7.50  & 8.00  & 9.00  & 8.00  & 8.00 \\
    \bottomrule
    \bottomrule
    \end{tabular}%
} 
\vspace{-0.2cm}
  \label{tab:metrics_synthetic_graphs}%
\end{table}%

\vspace{-0.1cm}
To compare the metrics quantitatively, we measure the similarity between the metric and GNN (GCN and SGC) curves by Pearson correlation and Fréchet distance. Pearson correlation~\cite{Pearson1895Regression} measures linear correlation between two sets of data; Fréchet distance measures the similarity between two arbitrary curves and it can be approximately calculated by the discrete Fréchet distance\footnote{We use the Python implementation for the calculation of discrete Fréchet distance provided by~\cite{eiter1994computing}. The code is from \url{https://pypi.org/project/frechetdist/}.}~\cite{alt1995computing, efrat2002new}. A smaller distance value indicates a higher similarity.  We calculate the average ranking of the homophily metrics \wrt{} graph generation methods and similarity metrics. From the results in Table~\ref{tab:metrics_synthetic_graphs} we can see that,
\begin{itemize}
   \item \textbf{Classic Metrics Are Still Strong} Although some new proposed metrics can reveal the rebounding phenomenon in extremely low homophily area, \eg{} $\text{H}_\text{agg}$, the old homophily metrics, \eg{}$\text{H}_\text{edge}, \text{H}_\text{node}, \text{H}_\text{class}$, still show very strong overall performance among different scenarios. On the other hand, many new metrics are unstable and exhibit unsatisfactory results in some cases, \eg{} $\text{H}_\text{agg}$, $\text{LI}$ and the hypothesis testing based metrics. 
   \item \textbf{Results Highly Depends on Similarity Measurements} For example, in RG generated graphs, the hypothesis testing based metrics show strong results with Pearson correlation but poor results with Fréchet distance. This is because Pearson correlation does not consider the actual spatial distance between points, but Fréchet distance considers spatial arrangement of points. The hypothesis testing based metrics are mainly designed to find out the threshold value (p-value) for good and bad graphs. Therefore, they might only capture the correlation instead of the spatial position of GNN curves. In the future, if people aim to design metrics with threshold values, we suggest using Pearson correlation as similarity measurement.
   \item \textbf{Discrepancy Between Different Synthetic Graphs} Given the same similarity measurement, the average ranking on different synthetic graphs can give quite different conclusions. For example, with Pearson correlation, $\text{KR}_\text{L}$ rank the best in RG generated graphs, but rank the worst in PA generated graphs. Similar problem happens to $\text{H}_\text{agg}, \text{LI}$ and $\text{GNB}$. In the future, to get a more reliable conclusion, we suggest people to take a more comprehensive view based on results from all the three synthetic graphs instead on just one of them.
\end{itemize}

\vspace{-0.3cm}
\section{Conclusion and Limitation}
\vspace{-0.2cm}
In this paper, we reveal and overcome three pitfalls in the model and metric evaluation for heterophilic graph representation learning: 1) lack of hyperparameter tuning; 2) insufficient model evaluation on the real challenging heterophilic datasets; 3) absence of quantitative evaluation benchmark for homophily metrics on synthetic graphs. We fine-tune the baseline models to discover three different types of heterophilic graphs among $27$ mostly used benchmark datasets, \ie{} malignant, benign and ambiguous heterophily datasets. We identify the real challenging tasks and reassess $11$ popular SOTA models with fine-tuned hyperparameters. At last, we conduct quantitative evaluation for $11$ popular metrics based on Pearson correlation and Fréchet distance and analyze performance. 

In the future, more research on ambiguous heterophily datasets is necessary for understanding the synergy of graph structure and model nonlinearity. A corresponding homophily metrics can be designed based on such synergy.

\clearpage

\bibliography{reference}
\bibliographystyle{abbrv}

\newpage
\section*{NeurIPS Paper Checklist}

The checklist is designed to encourage best practices for responsible machine learning research, addressing issues of reproducibility, transparency, research ethics, and societal impact. Do not remove the checklist: {\bf The papers not including the checklist will be desk rejected.} The checklist should follow the references and follow the (optional) supplemental material.  The checklist does NOT count towards the page
limit. 

Please read the checklist guidelines carefully for information on how to answer these questions. For each question in the checklist:
\begin{itemize}
    \item You should answer \answerYes{}, \answerNo{}, or \answerNA{}.
    \item \answerNA{} means either that the question is Not Applicable for that particular paper or the relevant information is Not Available.
    \item Please provide a short (1–2 sentence) justification right after your answer (even for NA). 
\end{itemize}

{\bf The checklist answers are an integral part of your paper submission.} They are visible to the reviewers, area chairs, senior area chairs, and ethics reviewers. You will be asked to also include it (after eventual revisions) with the final version of your paper, and its final version will be published with the paper.

The reviewers of your paper will be asked to use the checklist as one of the factors in their evaluation. While "\answerYes{}" is generally preferable to "\answerNo{}", it is perfectly acceptable to answer "\answerNo{}" provided a proper justification is given (e.g., "error bars are not reported because it would be too computationally expensive" or "we were unable to find the license for the dataset we used"). In general, answering "\answerNo{}" or "\answerNA{}" is not grounds for rejection. While the questions are phrased in a binary way, we acknowledge that the true answer is often more nuanced, so please just use your best judgment and write a justification to elaborate. All supporting evidence can appear either in the main paper or the supplemental material, provided in appendix. If you answer \answerYes{} to a question, in the justification please point to the section(s) where related material for the question can be found.

IMPORTANT, please:
\begin{itemize}
    \item {\bf Delete this instruction block, but keep the section heading ``NeurIPS Paper Checklist"},
    \item  {\bf Keep the checklist subsection headings, questions/answers and guidelines below.}
    \item {\bf Do not modify the questions and only use the provided macros for your answers}.
\end{itemize}


\begin{enumerate}

\item {\bf Claims}
    \item[] Question: Do the main claims made in the abstract and introduction accurately reflect the paper's contributions and scope?
    \item[] Answer: \answerYes{}{} 
    \item[] Justification: 
    \item[] Guidelines: 
    \begin{itemize}
        \item The answer NA means that the abstract and introduction do not include the claims made in the paper.
        \item The abstract and/or introduction should clearly state the claims made, including the contributions made in the paper and important assumptions and limitations. A No or NA answer to this question will not be perceived well by the reviewers. 
        \item The claims made should match theoretical and experimental results, and reflect how much the results can be expected to generalize to other settings. 
        \item It is fine to include aspirational goals as motivation as long as it is clear that these goals are not attained by the paper. 
    \end{itemize}

\item {\bf Limitations}
    \item[] Question: Does the paper discuss the limitations of the work performed by the authors?
    \item[] Answer: \answerYes{} 
    \item[] Justification: last section
    \item[] Guidelines:
    \begin{itemize}
        \item The answer NA means that the paper has no limitation while the answer No means that the paper has limitations, but those are not discussed in the paper. 
        \item The authors are encouraged to create a separate "Limitations" section in their paper.
        \item The paper should point out any strong assumptions and how robust the results are to violations of these assumptions (e.g., independence assumptions, noiseless settings, model well-specification, asymptotic approximations only holding locally). The authors should reflect on how these assumptions might be violated in practice and what the implications would be.
        \item The authors should reflect on the scope of the claims made, e.g., if the approach was only tested on a few datasets or with a few runs. In general, empirical results often depend on implicit assumptions, which should be articulated.
        \item The authors should reflect on the factors that influence the performance of the approach. For example, a facial recognition algorithm may perform poorly when image resolution is low or images are taken in low lighting. Or a speech-to-text system might not be used reliably to provide closed captions for online lectures because it fails to handle technical jargon.
        \item The authors should discuss the computational efficiency of the proposed algorithms and how they scale with dataset size.
        \item If applicable, the authors should discuss possible limitations of their approach to address problems of privacy and fairness.
        \item While the authors might fear that complete honesty about limitations might be used by reviewers as grounds for rejection, a worse outcome might be that reviewers discover limitations that aren't acknowledged in the paper. The authors should use their best judgment and recognize that individual actions in favor of transparency play an important role in developing norms that preserve the integrity of the community. Reviewers will be specifically instructed to not penalize honesty concerning limitations.
    \end{itemize}

\item {\bf Theory assumptions and proofs}
    \item[] Question: For each theoretical result, does the paper provide the full set of assumptions and a complete (and correct) proof?
    \item[] Answer: \answerYes{} 
    \item[] Justification:
    \item[] Guidelines:
    \begin{itemize}
        \item The answer NA means that the paper does not include theoretical results. 
        \item All the theorems, formulas, and proofs in the paper should be numbered and cross-referenced.
        \item All assumptions should be clearly stated or referenced in the statement of any theorems.
        \item The proofs can either appear in the main paper or the supplemental material, but if they appear in the supplemental material, the authors are encouraged to provide a short proof sketch to provide intuition. 
        \item Inversely, any informal proof provided in the core of the paper should be complemented by formal proofs provided in appendix or supplemental material.
        \item Theorems and Lemmas that the proof relies upon should be properly referenced. 
    \end{itemize}

    \item {\bf Experimental result reproducibility}
    \item[] Question: Does the paper fully disclose all the information needed to reproduce the main experimental results of the paper to the extent that it affects the main claims and/or conclusions of the paper (regardless of whether the code and data are provided or not)?
    \item[] Answer: \answerYes{} 
    \item[] Justification: 
    \item[] Guidelines:
    \begin{itemize}
        \item The answer NA means that the paper does not include experiments.
        \item If the paper includes experiments, a No answer to this question will not be perceived well by the reviewers: Making the paper reproducible is important, regardless of whether the code and data are provided or not.
        \item If the contribution is a dataset and/or model, the authors should describe the steps taken to make their results reproducible or verifiable. 
        \item Depending on the contribution, reproducibility can be accomplished in various ways. For example, if the contribution is a novel architecture, describing the architecture fully might suffice, or if the contribution is a specific model and empirical evaluation, it may be necessary to either make it possible for others to replicate the model with the same dataset, or provide access to the model. In general. releasing code and data is often one good way to accomplish this, but reproducibility can also be provided via detailed instructions for how to replicate the results, access to a hosted model (e.g., in the case of a large language model), releasing of a model checkpoint, or other means that are appropriate to the research performed.
        \item While NeurIPS does not require releasing code, the conference does require all submissions to provide some reasonable avenue for reproducibility, which may depend on the nature of the contribution. For example
        \begin{enumerate}
            \item If the contribution is primarily a new algorithm, the paper should make it clear how to reproduce that algorithm.
            \item If the contribution is primarily a new model architecture, the paper should describe the architecture clearly and fully.
            \item If the contribution is a new model (e.g., a large language model), then there should either be a way to access this model for reproducing the results or a way to reproduce the model (e.g., with an open-source dataset or instructions for how to construct the dataset).
            \item We recognize that reproducibility may be tricky in some cases, in which case authors are welcome to describe the particular way they provide for reproducibility. In the case of closed-source models, it may be that access to the model is limited in some way (e.g., to registered users), but it should be possible for other researchers to have some path to reproducing or verifying the results.
        \end{enumerate}
    \end{itemize}

\item {\bf Open access to data and code}
    \item[] Question: Does the paper provide open access to the data and code, with sufficient instructions to faithfully reproduce the main experimental results, as described in supplemental material?
    \item[] Answer: \answerYes{} 
    \item[] Justification: 
    \item[] Guidelines:
    \begin{itemize}
        \item The answer NA means that paper does not include experiments requiring code.
        \item Please see the NeurIPS code and data submission guidelines (\url{https://nips.cc/public/guides/CodeSubmissionPolicy}) for more details.
        \item While we encourage the release of code and data, we understand that this might not be possible, so “No” is an acceptable answer. Papers cannot be rejected simply for not including code, unless this is central to the contribution (e.g., for a new open-source benchmark).
        \item The instructions should contain the exact command and environment needed to run to reproduce the results. See the NeurIPS code and data submission guidelines (\url{https://nips.cc/public/guides/CodeSubmissionPolicy}) for more details.
        \item The authors should provide instructions on data access and preparation, including how to access the raw data, preprocessed data, intermediate data, and generated data, etc.
        \item The authors should provide scripts to reproduce all experimental results for the new proposed method and baselines. If only a subset of experiments are reproducible, they should state which ones are omitted from the script and why.
        \item At submission time, to preserve anonymity, the authors should release anonymized versions (if applicable).
        \item Providing as much information as possible in supplemental material (appended to the paper) is recommended, but including URLs to data and code is permitted.
    \end{itemize}

\item {\bf Experimental setting/details}
    \item[] Question: Does the paper specify all the training and test details (e.g., data splits, hyperparameters, how they were chosen, type of optimizer, etc.) necessary to understand the results?
    \item[] Answer: \answerYes{}
    \item[] Justification: 
    \item[] Guidelines:
    \begin{itemize}
        \item The answer NA means that the paper does not include experiments.
        \item The experimental setting should be presented in the core of the paper to a level of detail that is necessary to appreciate the results and make sense of them.
        \item The full details can be provided either with the code, in appendix, or as supplemental material.
    \end{itemize}

\item {\bf Experiment statistical significance}
    \item[] Question: Does the paper report error bars suitably and correctly defined or other appropriate information about the statistical significance of the experiments?
    \item[] Answer: \answerYes{} 
    \item[] Justification:
    \item[] Guidelines:
    \begin{itemize}
        \item The answer NA means that the paper does not include experiments.
        \item The authors should answer "Yes" if the results are accompanied by error bars, confidence intervals, or statistical significance tests, at least for the experiments that support the main claims of the paper.
        \item The factors of variability that the error bars are capturing should be clearly stated (for example, train/test split, initialization, random drawing of some parameter, or overall run with given experimental conditions).
        \item The method for calculating the error bars should be explained (closed form formula, call to a library function, bootstrap, etc.)
        \item The assumptions made should be given (e.g., Normally distributed errors).
        \item It should be clear whether the error bar is the standard deviation or the standard error of the mean.
        \item It is OK to report 1-sigma error bars, but one should state it. The authors should preferably report a 2-sigma error bar than state that they have a 96\% CI, if the hypothesis of Normality of errors is not verified.
        \item For asymmetric distributions, the authors should be careful not to show in tables or figures symmetric error bars that would yield results that are out of range (e.g. negative error rates).
        \item If error bars are reported in tables or plots, The authors should explain in the text how they were calculated and reference the corresponding figures or tables in the text.
    \end{itemize}

\item {\bf Experiments compute resources}
    \item[] Question: For each experiment, does the paper provide sufficient information on the computer resources (type of compute workers, memory, time of execution) needed to reproduce the experiments?
    \item[] Answer: \answerYes{} 
    \item[] Justification: We put  it in the main paper
    \item[] Guidelines:
    \begin{itemize}
        \item The answer NA means that the paper does not include experiments.
        \item The paper should indicate the type of compute workers CPU or GPU, internal cluster, or cloud provider, including relevant memory and storage.
        \item The paper should provide the amount of compute required for each of the individual experimental runs as well as estimate the total compute. 
        \item The paper should disclose whether the full research project required more compute than the experiments reported in the paper (e.g., preliminary or failed experiments that didn't make it into the paper). 
    \end{itemize}
    
\item {\bf Code of ethics}
    \item[] Question: Does the research conducted in the paper conform, in every respect, with the NeurIPS Code of Ethics \url{https://neurips.cc/public/EthicsGuidelines}?
    \item[] Answer: \answerYes{}
    \item[] Justification: 
    \item[] Guidelines:
    \begin{itemize}
        \item The answer NA means that the authors have not reviewed the NeurIPS Code of Ethics.
        \item If the authors answer No, they should explain the special circumstances that require a deviation from the Code of Ethics.
        \item The authors should make sure to preserve anonymity (e.g., if there is a special consideration due to laws or regulations in their jurisdiction).
    \end{itemize}

\item {\bf Broader impacts}
    \item[] Question: Does the paper discuss both potential positive societal impacts and negative societal impacts of the work performed?
    \item[] Answer: \answerNA{}{} 
    \item[] Justification: We have carefully reevaluated our project and we do  not have such negative impact.
    \item[] Guidelines:
    \begin{itemize}
        \item The answer NA means that there is no societal impact of the work performed.
        \item If the authors answer NA or No, they should explain why their work has no societal impact or why the paper does not address societal impact.
        \item Examples of negative societal impacts include potential malicious or unintended uses (e.g., disinformation, generating fake profiles, surveillance), fairness considerations (e.g., deployment of technologies that could make decisions that unfairly impact specific groups), privacy considerations, and security considerations.
        \item The conference expects that many papers will be foundational research and not tied to particular applications, let alone deployments. However, if there is a direct path to any negative applications, the authors should point it out. For example, it is legitimate to point out that an improvement in the quality of generative models could be used to generate deepfakes for disinformation. On the other hand, it is not needed to point out that a generic algorithm for optimizing neural networks could enable people to train models that generate Deepfakes faster.
        \item The authors should consider possible harms that could arise when the technology is being used as intended and functioning correctly, harms that could arise when the technology is being used as intended but gives incorrect results, and harms following from (intentional or unintentional) misuse of the technology.
        \item If there are negative societal impacts, the authors could also discuss possible mitigation strategies (e.g., gated release of models, providing defenses in addition to attacks, mechanisms for monitoring misuse, mechanisms to monitor how a system learns from feedback over time, improving the efficiency and accessibility of ML).
    \end{itemize}
    
\item {\bf Safeguards}
    \item[] Question: Does the paper describe safeguards that have been put in place for responsible release of data or models that have a high risk for misuse (e.g., pretrained language models, image generators, or scraped datasets)?
    \item[] Answer: \answerNA{} 
    \item[] Justification: We have carefully reevaluated our project and we do  not have such risk.
    \item[] Guidelines:
    \begin{itemize}
        \item The answer NA means that the paper poses no such risks.
        \item Released models that have a high risk for misuse or dual-use should be released with necessary safeguards to allow for controlled use of the model, for example by requiring that users adhere to usage guidelines or restrictions to access the model or implementing safety filters. 
        \item Datasets that have been scraped from the Internet could pose safety risks. The authors should describe how they avoided releasing unsafe images.
        \item We recognize that providing effective safeguards is challenging, and many papers do not require this, but we encourage authors to take this into account and make a best faith effort.
    \end{itemize}

\item {\bf Licenses for existing assets}
    \item[] Question: Are the creators or original owners of assets (e.g., code, data, models), used in the paper, properly credited and are the license and terms of use explicitly mentioned and properly respected?
    \item[] Answer: \answerYes{} 
    \item[] Justification: 
    \item[] Guidelines:
    \begin{itemize}
        \item The answer NA means that the paper does not use existing assets.
        \item The authors should cite the original paper that produced the code package or dataset.
        \item The authors should state which version of the asset is used and, if possible, include a URL.
        \item The name of the license (e.g., CC-BY 4.0) should be included for each asset.
        \item For scraped data from a particular source (e.g., website), the copyright and terms of service of that source should be provided.
        \item If assets are released, the license, copyright information, and terms of use in the package should be provided. For popular datasets, \url{paperswithcode.com/datasets} has curated licenses for some datasets. Their licensing guide can help determine the license of a dataset.
        \item For existing datasets that are re-packaged, both the original license and the license of the derived asset (if it has changed) should be provided.
        \item If this information is not available online, the authors are encouraged to reach out to the asset's creators.
    \end{itemize}

\item {\bf New assets}
    \item[] Question: Are new assets introduced in the paper well documented and is the documentation provided alongside the assets?
    \item[] Answer: \answerNA{} 
    \item[] Justification: 
    \item[] Guidelines:
    \begin{itemize}
        \item The answer NA means that the paper does not release new assets.
        \item Researchers should communicate the details of the dataset/code/model as part of their submissions via structured templates. This includes details about training, license, limitations, etc. 
        \item The paper should discuss whether and how consent was obtained from people whose asset is used.
        \item At submission time, remember to anonymize your assets (if applicable). You can either create an anonymized URL or include an anonymized zip file.
    \end{itemize}

\item {\bf Crowdsourcing and research with human subjects}
    \item[] Question: For crowdsourcing experiments and research with human subjects, does the paper include the full text of instructions given to participants and screenshots, if applicable, as well as details about compensation (if any)? 
    \item[] Answer: \answerNA{} 
    \item[] Justification: We do not have such problem
    \item[] Guidelines:
    \begin{itemize}
        \item The answer NA means that the paper does not involve crowdsourcing nor research with human subjects.
        \item Including this information in the supplemental material is fine, but if the main contribution of the paper involves human subjects, then as much detail as possible should be included in the main paper. 
        \item According to the NeurIPS Code of Ethics, workers involved in data collection, curation, or other labor should be paid at least the minimum wage in the country of the data collector. 
    \end{itemize}

\item {\bf Institutional review board (IRB) approvals or equivalent for research with human subjects}
    \item[] Question: Does the paper describe potential risks incurred by study participants, whether such risks were disclosed to the subjects, and whether Institutional Review Board (IRB) approvals (or an equivalent approval/review based on the requirements of your country or institution) were obtained?
    \item[] Answer: \answerYes{} 
    \item[] Justification: We have carefully reevaluated our project and we do  not have such risk.
    \item[] Guidelines:
    \begin{itemize}
        \item The answer NA means that the paper does not involve crowdsourcing nor research with human subjects.
        \item Depending on the country in which research is conducted, IRB approval (or equivalent) may be required for any human subjects research. If you obtained IRB approval, you should clearly state this in the paper. 
        \item We recognize that the procedures for this may vary significantly between institutions and locations, and we expect authors to adhere to the NeurIPS Code of Ethics and the guidelines for their institution. 
        \item For initial submissions, do not include any information that would break anonymity (if applicable), such as the institution conducting the review.
    \end{itemize}

\item {\bf Declaration of LLM usage}
    \item[] Question: Does the paper describe the usage of LLMs if it is an important, original, or non-standard component of the core methods in this research? Note that if the LLM is used only for writing, editing, or formatting purposes and does not impact the core methodology, scientific rigorousness, or originality of the research, declaration is not required.
    \item[] Answer: \answerNA{} 
    \item[] Justification: LLM  only for  grammar check.
    \item[] Guidelines: 
    \begin{itemize}
        \item The answer NA means that the core method development in this research does not involve LLMs as any important, original, or non-standard components.
        \item Please refer to our LLM policy (\url{https://neurips.cc/Conferences/2025/LLM}) for what should or should not be described.
    \end{itemize}

\end{enumerate}

\clearpage


\appendix

\section{Homophily Metrics}
\label{appendix:metrics_on_homogeneous_graphs}
There are mainly four ways to define the metrics that describe the relations among node labels, features and graph structure to predict whether graph-aware models can outperform their coupled graph-agnostic counterparts.

\paragraph{Graph-Label Consistency} Four commonly used homophily metrics based on the consistency between node labels and graph structures are edge homophily~\cite{abu2019mixhop,zhu2020beyond}, node homophily~\cite{pei2020geom}, class homophily~\cite{lim2021new} and adjusted homophily~\cite{platonov2023characterizing}
defined as follows:
\begin{equation}
\begin{aligned}
& \text{H}_\text{edge}(\mathcal{G}) = \frac{\big|\{e_{uv} \mid e_{uv}\in \mathcal{E}, y_{u} = y_{v}\}\big|}{|\mathcal{E}|}; \text{H}_\text{node}(\mathcal{G}) = \frac{1}{|\mathcal{V}|} \sum_{v \in \mathcal{V}} \frac{\big|\{u \mid u \in \mathcal{N}_v, y_{u} = y_{v}\} \big|}{d_v}; \\
& \text{H}_\text{class}(\mathcal{G}) \!=\! \frac{1}{C\!-\!1} \sum_{k=1}^{C}\bigg[h_{k}
    \!-\! \frac{\big|\{v \!\mid\! Y_{v,k} \!=\! 1 \}\big|}{N}\bigg]_{+}, \text{ where } h_{k}\! =\! \frac{\sum_{v \in \mathcal{V}, Y_{v,k}\! =\! 1} \big|\{u \!\mid\!  u \in \mathcal{N}_v, y_{u}\!=\! y_{v}\}\big| }{\sum_{v \in \{v|Y_{v,k}=1\}} d_{v}};  \\
& \text{H}_\text{adj}(\mathcal{G}) = \frac{\text{H}_\text{edge} - \sum_{c=1}^C \bar{p}_c^2}{1-\sum_{c=1}^C \bar{p}_c^2}, \text{ where } \bar{p}_c = \frac{\sum_{v: y_v=c} d_v}{2|\mathcal{E}|} \\
\end{aligned}
\end{equation}
where $[a]_{+}=\max (a, 0)$, $h_{k}$ is the class-wise homophily metric~\cite{lim2021new}.

Note that $\text{H}_\text{edge}(\mathcal{G})$ measures the proportion of edges that connect two nodes in the same class; $\text{H}_\text{node}(\mathcal{G})$ evaluates the average proportion of edge-label consistency of all nodes; $\text{H}_\text{class}(\mathcal{G})$ tries to avoid sensitivity to imbalanced classes, which can make $\text{H}_\text{edge}(\mathcal{G})$ misleadingly large; $\text{H}_\text{adj}(\mathcal{G})$ is constructed to satisfy maximal agreement and constant baseline properties. The above definitions are all based on the \textbf{linear feature-independent graph-label consistency}. The inconsistency relation indicated by a small metric value implies that the graph structure has a negative effect on the performance of GNNs. 

\paragraph{Similarity Based Metrics} Generalized edge homophily~\cite{jin2022raw} and aggregation homophily~\cite{luan2022revisiting} leverage similarity functions to define the metrics,
\begin{equation}
\begin{aligned}
& \resizebox{1\hsize}{!}{$\text{H}_\text{GE} (\mathcal{G}) = \frac{\sum\limits_{(i,j) \in \mathcal{E}} \text{cos}(\bm{x}_{i}, \bm{x}_{j})}{|\mathcal{E}|};\ \text{H}_{\text{agg}}(\mathcal{G}) =  \frac{1}{\left| \mathcal{V} \right|} \times \left| \left\{v \,\big| \, \mathrm{Mean}_u  \big( \{S(\hat{{A}},Y)_{v,u}^{y_{u} =y_{v}} \}\big) \geq  \mathrm{Mean}_u\big(\{S(\hat{{A}},Y)_{v,u}^{y_{u} \neq y_{v}}  \} \big) \right\} \right|$} \\
\end{aligned}
\end{equation}
where $\mathrm{Mean}_u\left(\{\cdot\}\right)$ takes the average over $u$ of a given multiset of values or variables and $S(\hat{{A}},Y)=\hat{{A}}Y(\hat{{A}}Y)^\top$ is the post-aggregation node similarity matrix. These two metrics are feature-dependent.

$\text{H}_\text{GE} (\mathcal{G})$ generalizes $\text{H}_\text{edge}(\mathcal{G})$ to the the cosine similarities between node features; $\text{H}_{\text{agg}}(\mathcal{G})$ measures the proportion of nodes $v\in\mathcal{V}$ as which the average $S(\hat{{A}},Y)$ weights on the set of nodes in the same class (including $v$) is larger than that in other classes. They are both feature-dependent metrics.

\paragraph{Neighborhood Identifiability/Informativeness}  Label informativeness ~\cite{platonov2023characterizing} and neighborhood identifiability~\cite{chen2023exploiting} leverage the neighborhood distribution instead of pairwise comparison to define the metrics,

\begin{equation}
\begin{aligned}
\resizebox{1\hsize}{!}{$ \text{LI} = -\frac{\sum_{c_1, c_2} p_{c_1, c_2 } \log \frac{p_{c_1, c_2}}{\bar{p}_{c_1} \bar{p}_{c_2}}}{\sum_c \bar{p}_c \log \bar{p}_c};\ {H}_{\text {neighbor }}(\mathcal{G}) = \sum_{k=1}^C \frac{n_k}{N} \text{H}_{\text {neighbor }}^k, \text{ where } {H}_{\text {neighbor }}^k = \frac{-\sum_{i=1}^C \tilde{\sigma}_i^k \log \left( \tilde{\sigma}_i^k \right)}{\log (C)}.$}
\end{aligned}
\end{equation}

where $p_{c_1, c_2} = \sum_{(u, v) \in \mathcal{E}} \frac{\bm{1} \left\{y_u=c_1, y_v=c_2 \right\}}{2|\mathcal{E}|}, c_1, c_2 \in \{1,\dots,C\}$; $A^k \in \mathbb{R}^{n_k \times C}$ is a class-level neighborhood label distribution matrix for each class $k$, where $k=1, \ldots, C$ for different classes and $n_k$ indicates the number of nodes with the label $k$, $(A^k)_{i,c}$ is the proportion of the neighbors of node $i$ belonging to class $c$, $\sigma_1^k, \sigma_2^k, \ldots, \sigma_C^k$ denote singular values of $A^k$, they are normalized to $\sum_{c=1}^C \tilde{\sigma}_c^k=1$, $c=1, \ldots, C$ is the index of singular values.

$\text{LI}$ is to characterize different connectivity patterns by measuring the informativeness of the label of a neighbor for the label of a node; $\text{H}_{\text {neighbor}}(\mathcal{G})$ is a weighted sum of ${H}_{\text {neighbor}}^k(\mathcal{G})$, quantifying neighborhood identifiability through the entropy of the singular value distribution of $A^k$, which is a generalization of the von Neumann entropy in quantum statistical mechanics~\cite{bengtsson2017geometry} that measures the pureness/information of a quantum-mechanical system. This metric effectively measures the complexity/randomness of neighborhood distributions by indicating the number of vectors (or neighbor patterns) necessary to sufficiently describe the neighborhood label distribution matrix.
\paragraph{Hypothesis Testing Based Performance Metrics} Luan \etal~\cite{luan2023graph} proposed classifier-based performance metric (CPM),\footnote{Luan \etal~\cite{luan2023we} also conducted hypothesis testing to find out when to use GNNs for node classification, but what they tested was the differences between connected nodes and unconnected nodes instead of intra- and inter-class nodes and they did not propose a metric based on hypothesis testing.} which uses the p-value of hypothesis testing as the metric to measure the node distinguishability of the aggregated features compared with the original features.

They first randomly sample 500 labeled nodes from $\mathcal{V}$ and splits them into 60\%/40\% as "training" and "test" data. The original features $X$ and aggregated features $H$ of the sampled training and test nodes can be calculated and are then fed into a given classifier. The prediction accuracy on the test nodes will be computed directly with the feedforward method. This process will be repeated 100 times to get 100 samples of prediction accuracy for $X$ and $H = \hat{A}X$. Then, for the given classifier, they compute the p-value of the following hypothesis testing, 
\begin{equation}
\begin{aligned}
\resizebox{1\hsize}{!}{$\text{H}_0: \text{Acc}(\text{Classifier}(H)) \geq \text{Acc}(\text{Classifier}(X));\ \text{H}_1: \text{Acc}(\text{Classifier}(H)) < \text{Acc}(\text{Classifier}(X)) $}
\end{aligned}
\end{equation}
The p-value can provide a statistical threshold value, such as 0.05, to indicate whether $H$ is significantly better than $X$ for node classification. To capture the feature-based linear or non-linear information efficiently, Luan \etal{} choose Gaussian Naïve Bayes (GNB)~\cite{hastie2009elements} and Kernel Regression (KR) with Neural Network Gaussian Process (NNGP)~\cite{lee2018deep,arora2019exact, garriga2018deep,matthews2018gaussian} as the classifiers, which do not require iterative training.

Overall, $\text{H}_\text{adj}$ can assume negative values, while other metrics all fall within the range of $[0,1]$. Except for $\text{H}_{\text {neighbor}}(\mathcal{G})$, where a smaller value indicates more identifiable,\footnote{Note that we use $1-\text{H}_{\text {neighbor}}(\mathcal{G})$ for quantitative analysis in this paper.} the other metrics with a value closer to $1$ indicate strong homophily and suggest that the connected nodes tend to share the same label, implying that graph-aware models are more likely to outperform their coupled graph-agnostic model, and vice versa. 

$\text{H}_\text{edge}, \text{H}_\text{node}, \text{H}_\text{class}, \text{H}_\text{adj}$ and $\text{LI}$ are linear feature-independent metrics. $LI$ and ${H}_{\text{neighbor}}(\mathcal{G})$ are nonlinear feature-independent metrics. $\text{H}_\text{GE}$ and $\text{H}_{\text{agg}}$ are feature-dependent and measure the linear similarity between nodes. CPM is the first metric that can capture nonlinear feature-dependent information and provide accurate threshold values to indicate the superiority of graph-aware models. In Section~\ref{sec:synthetic_comparison}, we will introduce the approach for the comparison of the above metrics by synthetic graphs with different generation methods.\section{More Experimental Setups}

\subsection{Hyperparameter Searching Range}
\label{appendix:hyperparameter_searching_range}

For every models and datasets, we perform a grid search for learning rate $\in\{{0.01,0.05,0.1}\}$, weight decay $\in\{0, 5e-7,5e-6, 1e-5, 5e-5, 1e-4, 5e-4, 1e-3, 5e-3, 1e-2\}$, dropout $\in\{ 0, 0.1,0.3, 0.5, 0.7\}$ with the Adam optimizer. We use hidden unit $=128$ for wiki-cooc, 512 for roman-empire, amazon-ratings,  minesweeper, tolokers, questions, Squirrel-filtered, Chameleon-filtered, and 64 for all the other datasets. These settings are used for GCN, SGC-1, MLP-2, MLP-1, and shared by other GNN models. Specific hyperparameters for are listed as follows 
\begin{itemize}
    \item GPRGNN: the weight is initialized  by their Personalized PageRank, 
    $\alpha \in \{ 0.1, 0.2,  0.5,  0.9 \}$ and $K=10$ power of the adjacency is used.
    \item BernNet: the propagation steps $K=10$.
    \item FAGCN: $\epsilon \in \{ 0.3, 0.4, 0.5\}$
    \item LINKX: the number of layers of $\text{MLP}_{\mathbf{A}}$ and $\text{MLP}_{\mathbf{X}}$ are in $ \{1, 2 \}$.
    \item ACM-GCN: ``structure\_info" $\in \{0,1\}$, ``variant" $\in \{0,1\}$, with ``ACM-GCN+" and ``ACM-GCN++".
    \item GBK-GNN: we set $\lambda = 30$ and use the model based on GraphSage.
    \item FSGNN: 3-hop configuration under ``all-feature" settings. 
    \item APPNP: $\alpha \in \{ 0.1, 0.2,  0.5,  0.9 \}$ and $K=10$ power of the adjacency is used.
    
\end{itemize}

\subsection{Hyperparameter Fine-Tuning for Fair and Reliable Comparison}
\vspace{-0.1cm}
\label{sec:hyperparameter_tunning}
\begin{table}[htbp]
  \centering
  \caption{Comparison of baseline models with (w) and without (w/o) hyperparameter tuning. The results are highlighted in \red{red} if hyperparameter tuning significantly improve the model performance. The un-tuned models use the hyperparameters provided in~\cite{platonov2022critical}.}
  \resizebox{1\hsize}{!}{
    \begin{tabular}{c|p{6em}c|p{6em}c|cc|p{6em}c}
    \toprule
    \toprule
    \multirow{2}[4]{*}{Datasets/Models} & \multicolumn{2}{c|}{GCN} & \multicolumn{2}{c|}{MLP-2} & \multicolumn{2}{c|}{SGC-1} & \multicolumn{2}{c}{MLP-1} \\
\cmidrule{2-9}          & \multicolumn{1}{c}{w/o} & with     & \multicolumn{1}{c}{w/o} & with     & w/o   & with     & \multicolumn{1}{c}{w/o} & with \\
    \midrule
    Squirrel-filtered & 30.35 $\pm$ 1.71 & \textcolor[rgb]{ 1,  0,  0}{37.33 $\pm$ 1.88} &  26.20 $\pm$ 1.46 & \textcolor[rgb]{ 1,  0,  0}{38.30 $\pm$ 1.22} & {30.81 $\pm$ 1.69} & \textcolor[rgb]{ 1,  0,  0}{37.54 $\pm$ 2.13} & 28.78 $\pm$ 1.38 & 30.14 $\pm$ 1.53 \\
    Chameleon-filtered & 39.39 $\pm$ 3.81 & 41.46 $\pm$ 3.42 & 29.24 $\pm$ 3.17 & \textcolor[rgb]{ 1,  0,  0}{38.06 $\pm$ 3.98} & 37.64 $\pm$ 2.95 & \textcolor[rgb]{ 1,  0,  0}{44.00 $\pm$ 3.10} & {29.54 $\pm$ 3.77} & \textcolor[rgb]{ 1,  0,  0}{35.72 $\pm$ 2.23} \\
     roman-empire  & 41.77 $\pm$ 0.51 & \textcolor[rgb]{ 1,  0,  0}{48.92 $\pm$ 0.46} & 65.14 $\pm$ 0.60 & 66.04 $\pm$ 0.71 &  28.48 $\pm$ 1.01 & \textcolor[rgb]{ 1,  0,  0}{44.60 $\pm$ 0.52} & 52.67 $\pm$ 1.41 & \textcolor[rgb]{ 1,  0,  0}{64.12 $\pm$ 0.61} \\
     amazon-ratings & 45.28 $\pm$ 0.77 & \textcolor[rgb]{ 1,  0,  0}{50.05 $\pm$ 0.67} &  43.13 $\pm$ 0.95 & \textcolor[rgb]{ 1,  0,  0}{49.55 $\pm$ 0.81} & 38.00 $\pm$ 0.64 & 40.69 $\pm$ 0.42 & 36.46 $\pm$ 0.58 & 38.60 $\pm$ 0.41 \\
     minesweeper  & 71.73 $\pm$ 1.09 & 72.34 $\pm$ 0.93 & 50.10 $\pm$ 0.84 & 50.92 $\pm$ 1.25 & 49.54 $\pm$ 18.79 & \textcolor[rgb]{ 1,  0,  0}{82.04 $\pm$ 0.77} & 49.88 $\pm$ 1.30 & 50.59 $\pm$ 0.83 \\
     tolokers & {63.74 $\pm$ 3.30} & \textcolor[rgb]{ 1,  0,  0}{77.44 $\pm$ 1.32} & 70.73 $\pm$ 1.07 & \textcolor[rgb]{ 1,  0,  0}{74.58 $\pm$ 0.75} & {46.91 $\pm$ 15.67} & \textcolor[rgb]{ 1,  0,  0}{73.80 $\pm$ 1.35} & {45.64 $\pm$ 11.00} & \textcolor[rgb]{ 1,  0,  0}{71.89 $\pm$ 0.82} \\
     questions & 55.21 $\pm$ 1.52 & \textcolor[rgb]{ 1,  0,  0}{72.72 $\pm$ 1.93} & 70.95 $\pm$ 1.20 & 69.97 $\pm$ 1.16 & 51.59 $\pm$ 3.97 & \textcolor[rgb]{ 1,  0,  0}{70.06 $\pm$ 0.92} & 51.70 $\pm$ 3.14 & \textcolor[rgb]{ 1,  0,  0}{70.33 $\pm$ 0.96} \\
    \bottomrule
    \bottomrule
    \end{tabular}%
    }\vspace{-0.2cm}
  \label{tab:hyperparameter_finetuning_comparison}%
\end{table}%

\vspace{-0.1cm}
To demonstrate the importance of hyperparameter fine-tuning, following~\cite{luan2023graph}, we first fine-tune two baseline GNNs, GCN~\cite{kipf2016classification} and 1-hop SGC (SGC-1)~\cite{wu2019simplifying}, and their coupled graph-agnostic models, MLP-2 and MLP-1~\footnote{Note that for fair evaluation, we remove all tricks in model architectures, such as residual connection and batch normalization, and only keep the original models for tests.} on the datasets where the baseline models are claimed to be quite robust to hyperparameter values~\cite{platonov2022critical}. From the experimental results shown in Table~\ref{tab:hyperparameter_finetuning_comparison}, we have identified a serious pitfall for model evaluation on heterophilic datasets, \ie{} there exits a huge discrepancy between the model performance with and without (w/o) hyperparameter fine-tuning, even on the 'hyperparameter-robust' datasets. In Table~\ref{tab:hyperparameter_finetuning_comparison}, we can see that in $19$ out of $28$ cases, hyperparameter fine-tuning can significantly improve model performance. This implies that a large amount of reported results in existing literature are potentially unreliable if there is no fine-tuning or the hyperparameter searching range is not large enough. This pitfall significantly hinders the fair model comparison and disrupt our way to discover the real challenging heterophilic datasets and really effective models.
(See Appendix~\ref{appendix:hyperparameter_searching_range} for our hyperparameter searching range.)

\vspace{-0.2cm}

\end{document}